\documentclass[a4paper, 10pt, conference]{ieeeconf}      
\usepackage{FG2024}
\usepackage{comment}
\usepackage{graphicx}
\usepackage{amsmath}
\usepackage{amsfonts}

\usepackage{times}
\usepackage{epsfig}

\usepackage{graphicx}

\makeatletter
\@namedef{ver@everyshi.sty}{}
\makeatother
\usepackage{tikz}
\usepackage{comment}
\usepackage{amsmath,amssymb} %
\usepackage{wrapfig} %
\usepackage{color}

\usepackage{mathtools}
\usepackage[accsupp]{axessibility}  %

\usepackage[utf8]{inputenc} %
\usepackage[T1]{fontenc}    %
\usepackage{xr-hyper}
\usepackage[pagebackref=true,breaklinks=true,letterpaper=true,colorlinks,bookmarks=false]{hyperref}
\usepackage{url}            %
\usepackage{booktabs}       %
\usepackage{xcolor,colortbl}         %
\usepackage{caption}

\usepackage{multirow}
\let\labelindent\relax
\usepackage{enumitem}

\usepackage[normalem]{ulem}

\usepackage{subcaption}
\usepackage{float}
\usepackage{lscape}  

\usepackage{xspace}
\usepackage{setspace}
\usepackage{pifont}
\usepackage{comment}
\usepackage{bm}

\IEEEoverridecommandlockouts                              
\def\FGPaperID{9} 
\usepackage{subcaption}
\title{\LARGE \bf
CasCalib: Cascaded Calibration for Motion Capture from Sparse Unsynchronized Cameras
}

\author{\parbox{16cm}{\centering
    {\large Huibert Kwakernaak$^1$ and Pradeep Misra$^2$}\\
    {\normalsize
    $^1$ Faculty of Electrical Engineering, Mathematics and Computer Science, University of Twente, Enschede, The Netherlands\\
    $^2$ Department of Electrical Engineering, Wright State University, Dayton, USA}}
    \thanks{This work was not supported by any organization}
}

\usepackage{fancyhdr}
\renewcommand{\headrulewidth}{0pt}
\ifFGfinal
\else

\author{James Tang$^1$ \qquad Shashwat Suri$^1$ \qquad Daniel Ajisafe$^1$ \qquad Bastian Wandt$^2$ \qquad Helge Rhodin$^{1,3}$\\
$^1$The University of British Columbia \quad $^2$Linköping University
\quad $^3$Bielefeld University
}

\fi
\PassOptionsToPackage{table}{xcolor}

\definecolor{OliveGreen}{rgb}{0,0.6,0}
\usepackage{fancyhdr}

\PassOptionsToPackage{table}{xcolor}

\newcommand{\HR}[1]{{\color{pink}(Helge:#1)}}

\newcommand{\yes}{\textcolor{OliveGreen}{Yes}}
\newcommand{\no}{\textcolor{red}{No}}

\newcommand{\parag}[1]{\noindent\textbf{#1}}

\newcommand{\vg}{\mathbf{g}}

\newcommand{\vn}{\mathbf{n}}
\newcommand{\vo}{\mathbf{o}}
\newcommand{\vp}{\mathbf{p}}
\newcommand{\vq}{\mathbf{q}}

\newcommand{\vx}{\mathbf{x}}

\newcommand{\mK}{\mathbf{K}}

\newcommand{\mM}{\mathbf{M}}

\newcommand{\mR}{\mathbf{R}}

\newcommand{\mT}{\mathbf{T}}

\definecolor{Gray}{gray}{0.85}

\definecolor{DeepGreen}{rgb}{0.15,0.60,0.15}

\begin{document}
\maketitle

\thispagestyle{fancy}
\renewcommand{\headrulewidth}{0pt}
\fancyhf{}
\fancyhead[C]{2024 18th International Conference on Automatic Face and Gesture Recognition (FG)}





\fancyfoot[L]{979-8-3503-9494-8/24/\$31.00 \copyright 2024 IEEE}

\pagestyle{empty}
\begin{abstract}
It is now possible to estimate 3D human pose from monocular images with off-the-shelf 3D pose estimators.
However, many practical applications require fine-grained absolute pose information for which multi-view cues and 
camera calibration are necessary.
Such multi-view recordings are laborious because they require manual calibration, and are expensive when using dedicated hardware. 
Our goal is full automation, which includes temporal synchronization, as well as intrinsic and extrinsic camera calibration. This is done by using persons in the scene as the calibration objects.
Existing methods either address only synchronization or calibration, assume one of the former as input, or have significant limitations. A common limitation is that they only consider single persons, which eases correspondence finding.
We attain this generality by partitioning the high-dimensional time and calibration space into a cascade of subspaces and introduce tailored algorithms to optimize each efficiently and robustly.  
The outcome is an easy-to-use, flexible, and robust motion capture toolbox that we release to enable scientific applications, which we demonstrate on diverse multi-view benchmarks. Project website: \href{https://github.com/tangytoby/CasCalib}{https://github.com/jamestang1998/CasCalib}.
\end{abstract}


\section{Introduction}
\label{sec:intro}

Computer vision has now reached the mainstream, enabling detailed 3D reconstructions from handheld video recordings with ubiquitous mobile phones. However, when aiming for the reconstruction of dynamic human performances, it still demands multiple cameras and manual calibration as single-view methods suffer from occlusions and depth ambiguities. 
Hence, applications in visual effects and medical studies, such as those in neuroscience studying motion deficits, still rely on dedicated motion capture studios. 
However, their manual calibration is cumbersome and error-prone, and their high cost renders them entirely inaccessible to smaller companies and labs with a non-technical background. 

To combat these issues, we propose an end-to-end framework for camera calibration and time-synchronization of a sparse-camera system. It directly serves the before-mentioned human pose and shape estimation tasks. In addition, once automatically estimated on full body pose, the calibration can be applied for reconstructing other objects, including fine-grained face reconstruction yielding more precise angle and shape estimates for embodied virtual reality, expression estimation, and gesture recognition.

\begin{figure}[t!]
\centering
\includegraphics[width=0.66\linewidth,trim={0.1cm 0 0.01cm 0},clip]{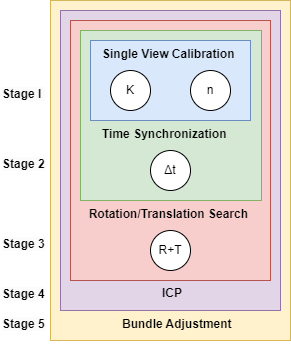}
\caption{\textbf{Cascaded calibration overview.} From top to bottom, we show how we break up the optimization problem into smaller subproblems by solving for a subset of the parameters at a time, with subsequent steps refining the earlier ones. The first step is the Single View Calibration step where we estimate the normal vector $\vn$ and the intrinsics $\mK$. Then, we estimate the time synchronization offset $\Delta t$. Finally, with the last three steps, we estimate and refine the rotation matrix $\mR$ and the translation $\mT$.}
\label{fig:overview} 
\end{figure}

Many approaches towards automating 3D capture exist, but only for specialized settings. Structure-from-motion techniques require either a continuous video stream of a static~\cite{schonberger2016structure} or slowly deforming object~\cite{bregler2000recovering}, or a dense array of cameras that have largely overlapping fields of view. These are popular for settings where there is only one camera or dozens of them, but nothing in between. We refer to this intermediate setting as the sparse camera case. For sparse camera setups, manual calibration with a checkerboard or other markers arranged in a predetermined two-dimensional pattern is the most common~\cite{chavdarova2018wildtrack}.
However, the calibration object needs to be carefully placed and repositioned by trained operators to ensure that the entire capture volume is covered, and multiple cameras see the calibration object.
Moreover, recalibration is required when cameras move ever so slightly, and fabricating calibration objects at very small or big scales, for example, meter scale for sports events, is challenging.

To achieve fully automated calibration, a promising fully automatic direction is to use humans or animals, which are usually present in the scene, as calibration objects. Fei et al.~\cite{fei2021single} calibrate the focal length and ground plane under the assumption that all people have roughly the same height and the ground is flat, but do not address synchronization and the multi-view case. Takahashi et al.~\cite{8575393} and Liu et al. ~\cite{https://doi.org/10.1049/cvi2.12130} subsequently use 2D keypoint detections in individual views to calibrate the extrinsic parameters of two or more cameras using classical fundamental matrix estimation, which requires seven or more correspondences across views. However, to establish the correspondence, they assume that cameras are synchronized, and only a single person is in view and visible from all cameras. In turn, Zhang et al. \cite{s21072464} establish temporal synchronization but assume intrinsic and extrinsic calibrated cameras, 
and still uses a single person. Xu et al. ~\cite{xu2021widebaseline} rely on reidentification, but synchronization is required and appearance matching is challenging when humans look alike, such as when a sports team dresses in the same uniform. To the best of our knowledge, there is no solution for unsynchronized sparse camera setups with a variable number of persons that are only partially visible.

We propose a cascaded calibration algorithm that breaks down the calibration of high-dimensional parameter space into subspaces that can be searched or optimized efficiently. For $N$ cameras, we solve for $N (4 \times 6 + 1)$ parameters which correspond to $4$ intrinsics (focal lengths and focal center), $6$ extrinsics (camera orientation and position), and $1$ temporal shift parameters. The outcome is a sequential process with steps having the cascading dependencies visualized in Figure~\ref{fig:overview}, with subsequent steps starting from preceding estimates that are further refined.
In the first stage, camera focal length (intrinsic) and their orientation with respect to the ground (extrinsic) are estimated similarly to \cite{fei2021single,tang2019}, independently in each view so that we don't need temporal synchronization across cameras. 
%
In the second stage, we estimate the temporal offset by taking the scalar distances of the ankles from the center of the ground plane in order to reduce the dimensions to one. Then we align the sequence temporally by searching for the frame offset that results in the minimum distance between the sequences.

In the third stage, we use these to reduce the 6D extrinsics problem to solving for 2D rotation and translation in the estimated ground planes, a 3D space optimized by least squares, and a greedy yet efficient search of the rotation on the ground plane. 
In the fourth stage, we refine the initialized extrinsic parameters alongside the additional temporal offset using ICP (\cite{besl1992method} and \cite{zhang1994iterative}) in a 5D space.
In the final stage, the entire 11D space is optimized using bundle adjustment \cite{10.5555/646271.685629}.

In summary, our main scientific contributions are:
\begin{itemize}
    \item The cascading decomposition of the calibration problem at an algorithmic level.
    \item Derivation of suitable objective functions and parameter representations for each stage, and the empirical validation of their hyperparameters.
\end{itemize}
A further contribution for computer vision practitioners is:
\begin{itemize}
    \item Source code, documentation, and curated environments to aid future work and deployment.
\end{itemize}

\section{Related Work}
We categorize methods for multi-view calibration based on human poses by their reconstruction methodology, such as optimization or deep learning; whether they can handle a single person or multiple persons; the required prior knowledge; and the output such as intrinsics, extrinsics, and temporal synchronization. Table \ref{tab:comparison} compares the capabilities and requirements of existing methods to ours.

\begin{table*}[t!]
\centering
  \resizebox{\linewidth}{!}{%
\begin{tabular}{|l|c|ccc|ccc|cc|}
\hline
Method                            & Framework & Finds Sync. & Finds Intrinsics & Finds Extrinsics & w/o GT Intrinsics & w/o GT Extrinsics & w/o GT Sync. & Multi-Person & Multi-view   \\
\hline
Lee \cite{9834083}                            & Deep         & \no  & \no  & \yes & \no  & \yes & \no  & \no  & \yes  \\
Qi Zhang \cite{zhang2022singleframe}          & Deep         & \yes & \no  & \no  & \no  & \no  & \yes & \yes & \yes \\
Chaoning Zhang \cite{Zhang_2020_WACV} & Deep & \no & \yes  & \yes  & \no & \no & \no  & N/A & \yes\\
Grabner \cite{grabner2019gp2c} & Deep & \no & \yes  & \yes  & \no & \no & \no  & N/A & \no\\
\hline
Jarved \cite{897380}                          & Optimization & \no  & \no  & \no  & \yes & \yes & \no  & \yes & \yes   \\
Xu \cite{xu2021widebaseline}                  & Optimization & \no  & \no  & \yes & \no  & \yes & \no  & \yes & \yes  \\
Troung \cite{s19224989}                       & Optimization & \no  & \no  & \yes & \no  & \yes & \no  & \yes & \yes   \\
Liu \cite{https://doi.org/10.1049/cvi2.12130} & Optimization & \no  & \yes & \yes & \yes & \yes & \no  & \no & \yes\\
Fei \cite{fei2021single}                      & Optimization & \no  & \yes & \yes & \yes & \yes & \yes & \yes  & \no\\
Takahashi \cite{8575393}                      & Optimization & \yes & \no  & \yes & \no  & \yes & \yes & \yes & \yes\\
Zhe Zhang \cite{s21072464}                    & Optimization & \yes & \no  & \no  & \no  & \no  & \yes & \no  & \no\\
Ours                    & Optimization & \yes & \yes  & \yes & \yes & \yes & \yes  & \yes & \yes\\
\hline
\end{tabular}}
\caption{\textbf{Comparison of related methods.} We summarize the differences, including which parameters they estimate and whether they require ground truth input. Only ours is able to calibrate and synchronize multi-person sequences.}
\label{tab:comparison}
\end{table*}

\textbf{Optimization:} These methods approach the calibration problem by obtaining keypoints of the poses from each camera, and then geometrically solving for the camera poses and optionally the 3D human pose.  
Liu et al. \cite{https://doi.org/10.1049/cvi2.12130} performs intrinsic and extrinsic calibration on videos containing a single person. Since the videos contain a single person and are temporally synchronized, they do not need to find multi-view correspondences between the cameras since the person in one camera will always correspond to the person in the other camera. First, they obtain 2D keypoints using OpenPose \cite{cao2019openpose}, then they use multi-view correspondences of a single person to triangulate the 3D keypoints. Finally, they reproject the 3D keypoints back to 2D image coordinates and they optimize the reprojection error between the cameras. Similar to our last processing step, the intrinsic and extrinsic parameters are further optimized using a RANSAC loop \cite{10.1145/358669.358692}, followed by bundle adjustment to optimize the camera poses. Takahashi et al.  \cite{8575393} calibrate extrinsic and intrinsic parameters in a similar way except they extend the method by using prior information about human poses such as bone length constraint and smooth motion constraints. However, neither Liu et al. \cite{https://doi.org/10.1049/cvi2.12130} nor Takahashi et al. \cite{8575393} solve the multi-person case, which requires obtaining multi-view correspondences. In addition to the requirement of multi-view correspondences, they also depend on temporal synchronization.

Many methods try to solve the multi-person case using person reidentification. Xu et al. \cite{xu2021widebaseline} use matches bounding boxes between views. This method performs relatively well on datasets where people wear distinct clothing, as such on the Terrace and Basketball sequences \cite{4359319}, but performs comparatively worse on their ConstructSite dataset where everyone is wearing the same uniform. Another method of matching is to take reprojections between each camera and match the closest pose in 3D to each other. This is used in several methods (\cite{s19224989,897380,s21072464}) and has the advantage that they do not depend on visual features on the persons such as clothing, since those could have high variation even on the same person, although combinatorically optimizing the matches can be expensive even with efficient algorithms like the Hungarian algorithm.

\textbf{Deep Learning:} These approaches solve the calibration problem by fully regressing the camera parameters or predicting intermediate estimates through training on a labeled dataset. Lee et al. \cite{9834083} 
use a 3D pose estimator and then use the predicted 3D poses as a 3D calibration object
to optimize the camera poses such that 3D poses match and re-project to 2D pose estimates.
This process alternates between optimizing the camera pose and intrinsics and optimizing the human pose. 
Zhang et al. \cite{Zhang_2020_WACV} perform camera calibration on pan tilt zoom (PTZ) cameras that rotate and zoom but do not translate. 
Their application requires an online estimation.
They take in as input two images from two views and use a three-part pipeline: a feature extractor based on a Siamese network architecture \cite{bromley1993signature}, feature matching using correlation, and a regression network,  to automatically estimate the intrinsic and extrinsic parameters, as well as the distortion parameters. Although this method can solve a more challenging problem, in the case where the cameras are not static, they require a large training set of over 100,000 image pairs. In addition, they do not handle the case where cameras are unsynchronized.
Grabner et al. \cite{grabner2019gp2c} approach the problem by extending the Faster/Mask R-CNN framework \cite{He_2017_ICCV}, 
by augmenting it with a focal length predictor and refining it using the reprojection error. However, despite the fact that their method gets 10 percent better focal lengths than the baseline, they use a very specific dataset of common objects as references such as chairs, sofas, and cars without occlusion.
In general, these objects may not be present. Additionally, all deep learning methods introduce the need for a training set and a training procedure, which may not generalize to real-world scenes.

\textbf{Synchronization:}
The previously described methods rely on the fact that the cameras are temporally synchronized with each other, which is an assumption that is often violated in practice. Besides methods such as hardware synchronization, which requires a pre-recording step to do, human pose can be used for temporal synchronization. In \cite{s21072464}, Zhang et al. find the temporal offset by minimizing error in epipolar lines from reprojecting 2D pose detections. However, this method is in turn limited by the need for the cameras to be calibrated. In \cite{s22228900}, Eichler et al. perform both camera synchronization and camera calibration using distances between joints from 3D pose detectors for synchronization, followed by a multi-view reconstruction for the extrinsic parameters. However, this method requires using a 3D pose detector, as well as assuming intrinsics are known.
Zhang et al. \cite{zhang2022singleframe} perform camera synchronization using a neural network architecture by warping one view to another, which does not assume a constant frame rate but intrinsic and extrinsic calibrations.

\textbf{Summary:}
Although these methods have been shown to be effective, they all require some sort of ground truth such as temporal synchronization, intrinsic, and extrinsic which they use to solve for the unknowns. We seek to use only 2D keypoint information without assuming calibration, synchronization, a training dataset, or person re-identification.

\section{Method}

\begin{figure*}[t!]
\centering
\includegraphics[width=0.95\linewidth,trim={0.01cm 0 0.1cm 0},clip]{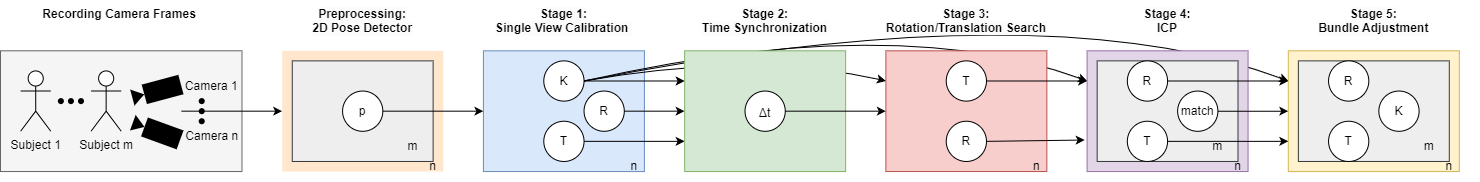}
  \caption{\textbf{System overview.} A fine-grained view of the five stages in Figure \ref{fig:overview}, including how detections of single persons in single views are treated independently in Stage I and jointly subsequently.
  Variables are in the plate notation, with $n$ the number of cameras and $m$ the number of people in the scene.}
\label{fig:overview_horizontal} 
\end{figure*}

To solve the camera calibration problem from multiple uncalibrated and unsynchronized views, we propose to break down the problem into several lower-dimensional problems. In a cascaded fashion, we start with a few variables that are solved globally and subsequently add details while reducing the range of the search space to stay practical.

We take in as input the 2D key point detections $\vp^{\text{img}} \in \mathbb{R}^2$ of the major human joints, such as the head, neck, and ankles.
Figure \ref{fig:overview} shows how the single view calibration, temporal search, rotation search, ICP, and bundle adjustment modules reduce the dimensionality of the problem, and Figure \ref{fig:overview_horizontal} shows how the variables are passed and refined between modules.

\subsubsection{Single View Geometric Calibration}

\begin{figure}
\centering
\includegraphics[width=0.6\linewidth,trim={0.1cm 0 0.01cm 0},clip]{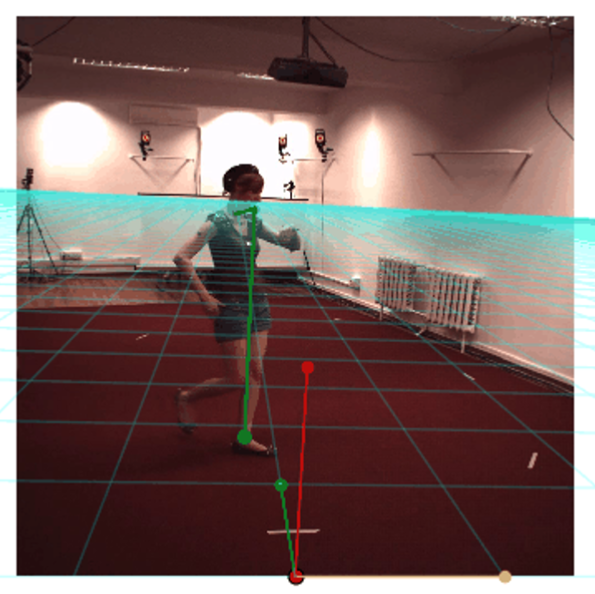}
    \caption{\textbf{2D Reconstruction.} Visual results for the single view calibration for Human3.6M Subject 1. The blue grid represents the ground plane predicted by our method with a coordinate axis defined at the bottom of the image. The green line from the ankle to the shoulders represents the ankle to shoulder keypoints.}
\label{fig:h36m_sub1} 
\end{figure}

In the single view calibration case, our goal is to find the intrinsic camera parameters $\mK$ of the projection transformation
\begin{equation}
\vp^\text{img} = \mK \vp^\text{cam},
\text{ where }
\mK = 
\begin{pmatrix}
f & 0 & o_1 \\
0 & f & o_2 \\
0 & 0 & 1
\end{pmatrix},
\label{eq:projection}
\end{equation}
mapping from 3D camera coordinates $\vp^\text{cam}$ to 2D image coordinates $\vp^\text{img}$. We estimate $\mK$, the ground plane position $\vg$, and orientation $\vn$ relative to the camera origin using a direct linear transform.
To make the estimation feasible with only $\vp^\text{img}$ as input, we assume that persons are standing up-right in some of the frames which makes them parallel to the ground plane normal vector and have a constant metric height $h$.
Furthermore, following the cascaded, coarse-to-fine principle, we fix the focal point $(o_1,o_2)$ to the image center and do not consider any relations across cameras as synchronization is missing. Note that these strong assumptions are lifted in later refinement stages.
\paragraph*{Direct Linear Transform} Using homogeneous coordinates, the ankle and shoulder positions of three or more people on a common ground plane are related by a linear system of equations that, when solved for its null space, reveal the sought-after camera parameters in closed form. The derivation is explained in the supplemental and is analogous to that in~\cite{fei2021single}. In practice, we apply RANSAC~\cite{10.1145/358669.358692} to filter out outliers and we select the largest inlier set. In order to determine which detections are inliers, we first reproject the ankle coordinates $\vp^{\text{img}}_{\text{ankle}}$ to 3D $\vp^{\text{cam}}_{\text{ankle}}$. Second, we add $h\vn$ to $\vp^{\text{cam}}_{\text{ankle}}$ to get the predicted $\vp^{\text{cam}}_{\text{shoulder}} = \vp_\text{ankle} + h\vn $. Finally, we reproject this point back to image coordinates using Eq.~\ref{eq:projection}. We use two metrics, shoulder pixel error and angle error to determine if it is an inlier. The pixel error is computed as the pixel Euclidean distance between the shoulder detection and the predicted shoulder. We normalize this by the 2D pixel height of the person. The angle error is computed by computing the angle between the vector from the $\vp^{\text{img}}_{\text{ankle}}$ to the $\vp^{\text{img}}_{\text{shoulder}}$ and the vector from $\vp^{\text{img}}_{\text{ankle}}$ to the predicted $\vp^{\text{img}}_{\text{shoulder}}$. An ankle-shoulder pair is considered an inlier if is less than both the angle and pixel threshold. For our experiments, we use an angle threshold of 2.86 degrees and a pixel threshold of 5 percent of the pixel height. Once we have the largest inlier set, we run our DLT method on the entire inlier set to get the final focal length and ground plane position and orientation. We show an example of our single view calibration algorithm in Figure \ref{fig:h36m_sub1}.

\paragraph*{Filtering} Since we have an assumption that the persons are standing straight up, we must filter out the non-standing poses in the scene. To determine this, for every 2D pose $\vp$, we measure the 2D angles between the vectors from shoulder to hip, hip to knee, and knee to ankle. We give the exact equations in the supplemental.

\begin{figure}
\centering
\includegraphics[width=0.8\linewidth,trim={0.1cm 0 0.01cm 0},clip]{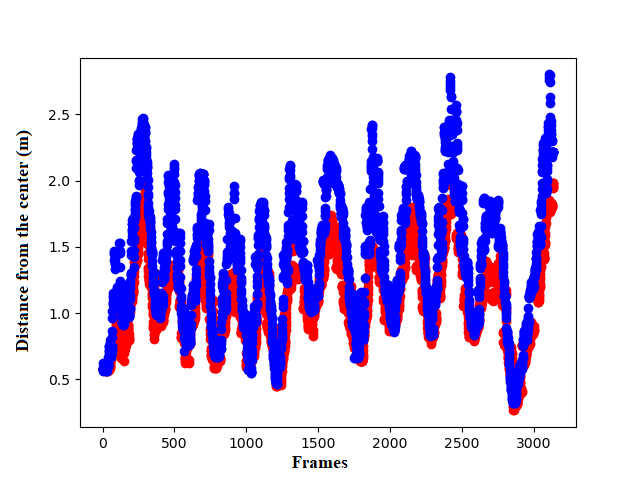}
  \caption{\textbf{Time synchronization.} Time synchronization results between ref (red) and sync (blue) sequences for subject 1 walking sequence in Human3.6M.}
\label{fig:sync_time} 
\end{figure}

\paragraph*{Relating multiple cameras through their ground planes}
The relation between individually calibrated cameras is unknown, but the estimated ground plane is shared.
We select one camera to be the reference camera, and use its plane coordinates as the world coordinate system. 
To simplify subsequent steps, we compute for each other camera the homography transformation from image coordinates $\vp^\text{img}$ to the estimated ground plane,
\begin{equation}
\vp^\text{plane} = [\mR^{\text{cam} \rightarrow \text{plane}} | \bm{\tau^\text{plane}}] \mK^{-1}\vp^\text{img}.
\end{equation}
Figure~\ref{fig:terrace_bird} shows the resulting birds-eye view of the ground plane with estimated person positions.
The plane normal vector $\vn$ is shared between all cameras and defines one column of $\mR^{\text{cam} \rightarrow \text{plane}}$.
For each camera, we derive the other 2 basis vectors by the backprojection of the 2D horizontal line to 3D as the new x-axis, and finally the cross product of the normal vector with the x-axis as the z-axis. 
The position 
$\bm{\tau}^\text{plane}$ is the back projection of the image center to the ground plane. 
This construction is intermittent. It remains to align the 2D position and orientation within the ground plane to fully determine the camera extrinsics, as well as to estimate the time shift. 

\begin{figure}[]
        \centering
        \includegraphics[width=0.75\linewidth, trim=0.1cm 0 0.01cm 0, clip]{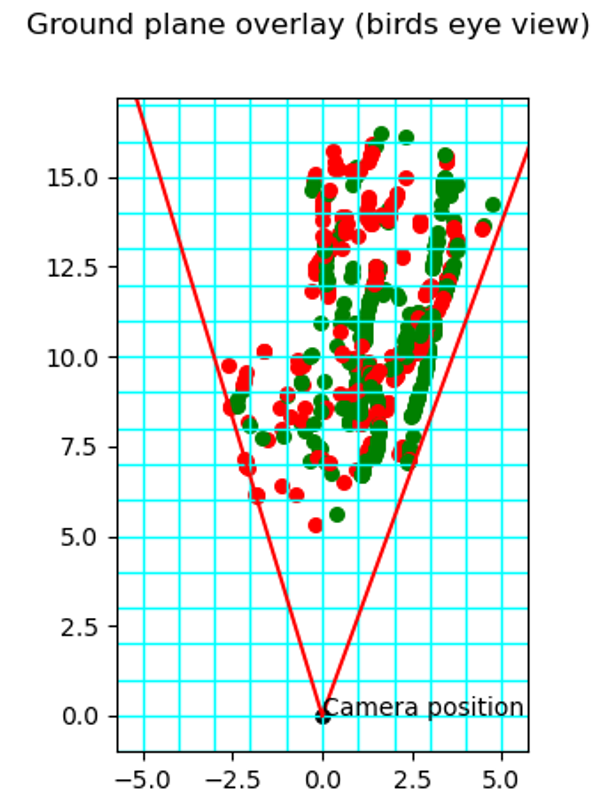}
        \label{fig:terrace_bird}

    \caption{\textbf{Ground plane view.} Bird-eye view of the ankles on the Terrace sequence, with inliers in green and outliers in red.}
    \label{fig:terrace_bird}
\end{figure}

\subsubsection{1D Temporal Search}
To support cameras starting or ending at different times, we model the time relationship pairwise between cameras as $t_{\text{ref}} = t_\text{sync} + \Delta t_\text{sync}$, a linear relationship  
between reference camera sequence $t_{\text{ref}}$ and $t_\text{sync}$ of the target camera.
To find the optimal translation $\Delta t_\text{sync}$, first, we project the detected ankle points onto the ground plane using $\mR^{\text{cam} \rightarrow \text{plane}}$ and shift them such that the mean of the reference set is the same as the mean of sync set. In order to get a signal that is time-sensitive but does not depend on the unknown camera extrinsics, we compute the distance $d$ from the center for each point. Since the search space is 1D, we can afford a brute-force search, with candidate offsets ranging from 0 to one-third of the length of the sync sequence.
Note that if there is more than one person in the frame, then that time step has more than one distance associated with it. We show an example of the temporal alignment in Figure \ref{fig:sync_time}.

\paragraph*{Search criteria}
The alignment is scored by
the absolute difference of $d_\text{sync}$ and $d_\text{ref}$
within the same time step. If there is more than one person in the frame, we compute an optimal matching 
using the Hungarian algorithm \cite{kuhn1955hungarian}. While shifting the curves temporally, we continue the endpoints of the curves by repeating the endpoint values. This helps prevent the curve from shifting too much since larger shifts lead to smaller overlap and hence larger uncertainty since we are not matching large amounts of the curve.

\paragraph*{Filtering} Since noisy detections could cause outlier points to appear on the ground plane, we remove outlier points on the ground plane using a density-based spatial clustering of applications with noise (DBSCAN) \cite{10.5555/3001460.3001507} to find the largest cluster of points on the ground plane, and then remove all the outlier points.

\subsubsection{2D Rotation Search}
\begin{figure}[]
    \centering
    \begin{subfigure}[t]{0.48\linewidth} 
        \includegraphics[width=\linewidth, trim=0.1cm 0 0.01cm 0, clip]{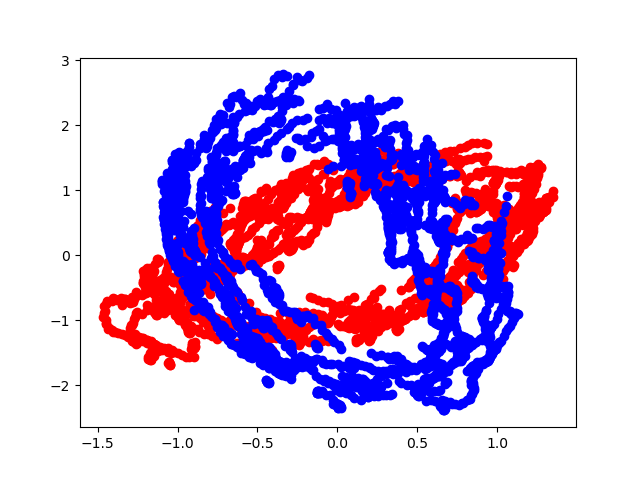}
        \caption{Initial orientation}
        \label{fig:rot_init}
    \end{subfigure}
    \hfill 
    \begin{subfigure}[t]{0.48\linewidth}
        \includegraphics[width=\linewidth, trim=0.1cm 0 0.01cm 0, clip]{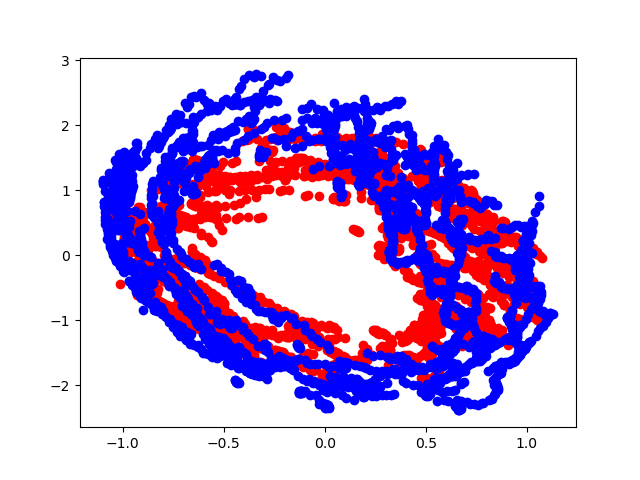}
        \caption{Best rotation}
        \label{fig:rot_best}
    \end{subfigure}
    \caption{\textbf{2D Rotation search.} Visual results for 2D rotation search on Human3.6M subject 1 with 2 cameras. Axes are in meters.}
    \label{fig:rotation_search}
\end{figure}

Once we match the videos temporally, we complete the extrinsic calibration between the cameras. First, we shift the means of the ankle positions in 2D plane coordinates from the \emph{sync} camera sequence to align with the reference camera's sequence. Then we search rotation angles from 0 to 360 degrees and apply the rotation to each sequence. For each camera i, we compute our error augmenting our detections $\vp^{\text{plane}}_i$ with the time step, $\hat{\vp}^{\text{plane}}_i = (x, y, t)$, and computing the distance between the 
closest
points in the point cloud. Note that no closed-form solution is possible, since correspondences between the two point clouds are unknown. We elaborate further on the equations in the supplemental. We show an example of the point cloud alignment process in Figure \ref{fig:rotation_search}. 

\subsubsection{Iterative Closest Point}

The previous section
yields a first estimate of all camera extrinsics by estimating the 2D plane rotation and position of cameras relative to each other that remained unknown in Step~1.
We refine that estimate using the Iterative Closest Point (ICP). 
We find the closest points by utilizing the previously estimated time synchronization to match the frames from the reference view to the synchronization view and also the same Hungarian matching process, when multiple persons are present. 
We then optimize the rotation and translation by minimizing the Euclidean distance between the 2D point clouds in the plane
Iteratively, we re-associate the points and repeat the process. Note that the initial 2D rotation search searches all angles between 0 and 360, making adjustments from the ICP small.

\subsubsection{Joint Camera Refinement (Bundle Adjustment)}

\begin{figure}[]
        \includegraphics[width=\linewidth, trim=0.1cm 0 0.01cm 0, clip]{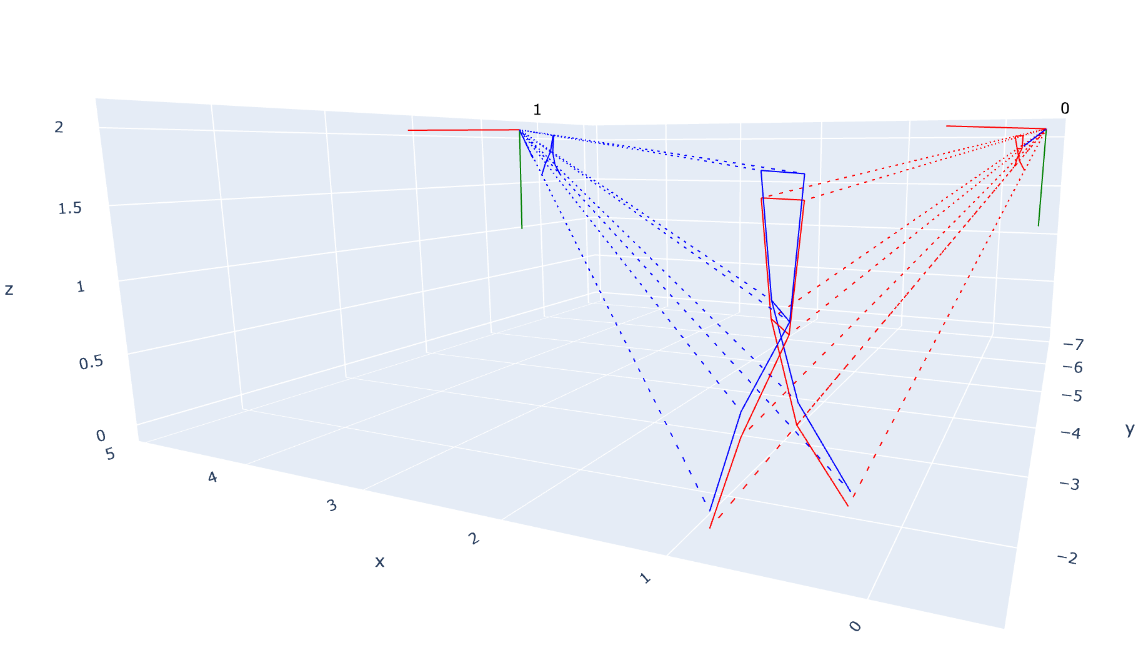}
        \label{fig:bundle_adjustment}

    \caption{\textbf{Bundle adjustment.} Visual 3D pose reconstruction for Human3.6M subject 1 with 2 cameras where the red and blue poses represent the closest points on each view ray}
    \label{fig:bundle_adjustment_all}
\end{figure}

Once we get the result from our ICP step, we pick the top $k$ poses with the highest confidence to use in the final bundle adjustment~\cite{10.5555/646271.685629} step that refines all calibration parameters jointly, by using the association of keypoints from the previous timesteps. 
As opposed to previous steps using ankle and shoulder, we can now incorporate all body part detections. 
Figure~\ref{fig:bundle_adjustment_all} shows this triangulation.
However, because the head and arm keypoints have a larger range of motion and are often self-occluded, we exclude them during the bundle adjustment. For camera pairs $i$ and $j$, we can define the relationship from 2D to 3D as a line using $\ell(k) = m_jk + \bm{\tau}_j$, with $k \in \mathbb{R}$ and 
\begin{equation}
m_j = \mR^{\text{plane} \rightarrow \text{world}}_j
(\mR^{\text{cam} \rightarrow \text{plane}}_j \mK_i^{-1} \textbf{P}^\text{img}_j - \bm{\tau}^{\text{cam} \rightarrow \text{plane}}_j).
\end{equation}
%
We then minimize the distance between pairs of lines by gradient descent.

\section{Experiments}

For our experiments, we test different components of the pipeline on a variety of datasets including Human3.6M \cite{h36m_pami}, EPFL Terrace and Laboratory \cite{4359319}, and vPTZ \cite{possegger12a}.
To aid comparisons with future work, we will make the code publicly available. As in Fei et al., we adopt HRNet \cite{sun2019deep}, as implemented by mmpose \cite{mmpose2020}, to obtain 2D keypoint detections and tracking.
For steps that require the results of previous steps, we run the previous parts of our pipeline to get the intermediate results to use as input to the next part of our pipeline.

\paragraph*{Metrics}
We use different metrics to evaluate different components of the camera calibration. Here is a brief breakdown of the metrics used:
\begin{itemize}[noitemsep,nolistsep]
    \item $\mathbf{Percent}$ $\mathbf{Focal}$ $\mathbf{Error}$ is used to measure the relative error in focal length and can be described as follows:
    \begin{equation}
    \mathbf{\%f} = \mathbf{\frac{f_{predicted} - f_{gt}}{f_{gt}}}
    \end{equation}
    \item $\mathbf{Relative}$ $\mathbf{Rotation}$ $\mathbf{and}$ $\mathbf{Translation}$: Since the coordinate system for the ground truth extrinsic is only uniquely defined up to a global rotation and translation, to evaluate the camera pose, we compute the relative rotation and translation from the reference camera to all the other cameras for both the predicted and the ground truth camera systems and then represent the error as follows
    \begin{align}
    \mathbf{^{\circ}} &= \mathbf{|^{\circ}_{\;\;rel} - ^{\circ}_{\;\;gt}|} \\
    \mathbf{m} &= \mathbf{|m_{rel} - m_{gt}|}
    \end{align}
\end{itemize}


\paragraph*{Datasets} We evaluate on the following datasets, each modeling a different setting in terms of the number of people and cameras, and the scale.
\begin{itemize}[noitemsep,nolistsep]
    \item $\mathbf{Human3.6M}$ \cite{h36m_pami} contains 4 cameras temporally synchronized and calibrated cameras that are recorded on a variety of subjects and actions. For our experiments, we use subjects 1,5,6,7,8,9,11 and use the walking action. Each video only contains one subject at a time. 
    \item $\mathbf{Terrace}$ $\mathbf{and}$ $\mathbf{Laboratory}$ \cite{4359319} from EPFL multi-camera pedestrian videos contains 4 temporally synchronized and calibrated cameras filming an outdoor scene with up to 7 subjects (Terrace) and another indoor scene with up to 4 subjects (Laboratory). Although the cameras are calibrated, the dataset has two versions of the calibrations. We show how we process these files in the supplemental. 
    \item $\mathbf{vPTZ}$ \cite{possegger12a} contains 3 outdoor scenes with numerous pedestrians with 4 cameras. Each video is filmed from a fairly high view and is representative of outdoor security camera footage.
\end{itemize}
\subsection{Single View Experiments}
\begin{figure*}[]
    \begin{subfigure}{0.23\linewidth} 
        \centering
        \includegraphics[width=\linewidth]{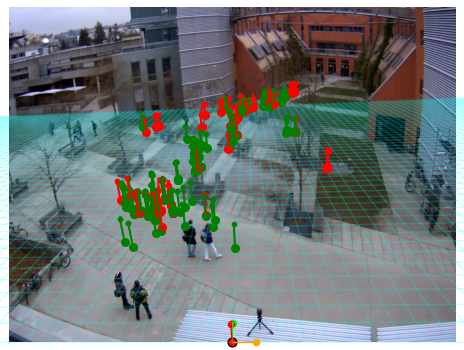}
        \subcaption{\textbf{vPTZ.} set1-cam-131 }
    \end{subfigure}%
    \hfill 
    \begin{subfigure}{0.23\linewidth}
        \centering
        \includegraphics[width=\linewidth]{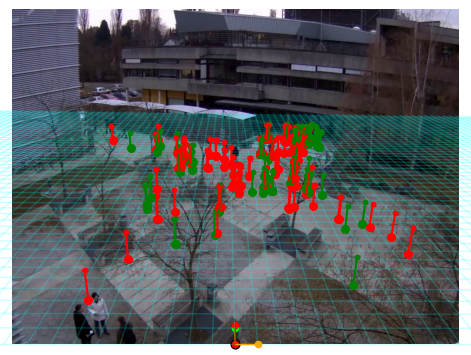}
        \subcaption{\textbf{vPTZ.} set1-cam-132}
    \end{subfigure}%
    \hfill
    \begin{subfigure}{0.23\linewidth}
        \centering
        \includegraphics[width=\linewidth]{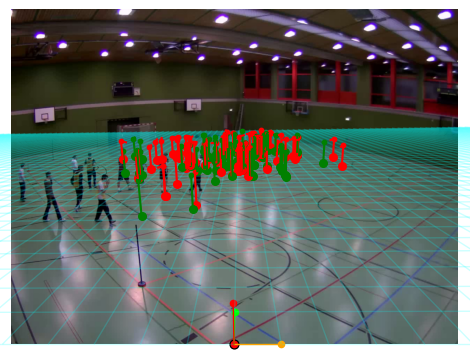}
        \subcaption{\textbf{vPTZ.} set2-cam-132}
    \end{subfigure}%
    \hfill
    \begin{subfigure}{0.23\linewidth}
        \centering
        \includegraphics[width=\linewidth]{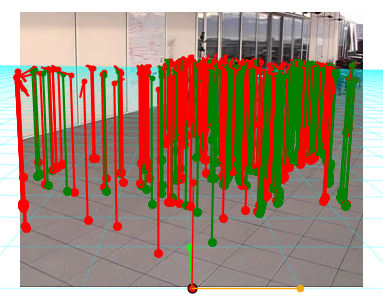}
        \subcaption{\textbf{Terrace.} terrace1-cam0}
    \end{subfigure}
    \caption{\textbf{Single view calibration results.} Visual results of our single view calibration algorithm from table \ref{tab:single_view_cal_results} with the predicted ground plane represented as a blue grid where every square is one meter. The green and red lines represent reprojected ankle and head detections from the entire sequence. Red represents outlier detections and green represents inlier detections.}
    \label{fig:single_view_cal_results}
\end{figure*}

To test the single view calibration stage, we test our method on vPTZ and the EPFL Terrace sequences. We compare against \cite{fei2021single} single view calibration method, and their results. In addition, we also compare against the other methods that Fei et al. implemented for testing, including \cite{liu2011}, \cite{liu2013}, \cite{Brouwers2016}, and \cite{tang2019}. We report our numerical results in Table \ref{tab:single_view_cal_results}
and visual results of the ground plane and shoulder reprojections in Figure \ref{fig:single_view_cal_results}. We find that our method is comparable to the related methods in the table, getting an average error of 11 percent of the ground truth focal length.
Although our method does not outperform the other methods, we deemed it to be a good enough reproduction to proceed with the main focus of this paper, as we did not have the code with additional implementation details and hyperparameter choices.

In addition to these experiments, we also evaluate our single view calibration on synthetic data, which is described in detail in the supplementary section. 

\begin{table}[ht]
\centering
\resizebox{1\linewidth}{!}{%
\begin{tabular}{|c|c|c|c|c|}
\hline
\multicolumn{1}{|c|}{\textbf{Method}} & \textbf{set1-cam-131 $\downarrow$} & \textbf{set1-cam-123 $\downarrow$} & \textbf{set2-cam-132 $\downarrow$} & \textbf{terrace1-cam0 $\downarrow$} \\
\hline
\cite{liu2011} & 1.00 & 29.00 & N/A & N/A \\
\hline
\cite{liu2013} & 2.00 & 19.00 & N/A & N/A \\
\hline
\cite{Brouwers2016} & N/A & 15.00 & 24.92 & 5.33 \\
\hline
\cite{tang2019} & N/A & 10.14 & 12.07 & 1.43 \\
\hline
\cite{fei2021single} & 4.70 & 0.35 & 10.74 & 2.51 \\
\hline
Ours & 11.31 & 0.66 & 18.36 & 13.18 \\
\hline
\end{tabular}%
}
\caption{\textbf{Single View Calibration Results.} Percent focal error on the vPTZ sequences.}
\label{tab:single_view_cal_results}
\end{table}

\subsection{Temporal Synchronization Experiments}
For comparing against existing synchronization methods, we test our method on Human3.6M. We randomly cut a section of each video that is half the length of the sequence for the Walking sequence for each subject. This is similar to the experiments that Zhang et al. \cite{s21072464} do except they only test their method when the true offset is 0, which is too simplistic. For our experiments, we shift the sequence with offsets 0, 50, 100, 150, and 200. We run the experiments using ground truth focal length and the focal length predicted by our method. We show that our method has a mean and median prediction that is close to the true offset, up to an error of 10 frames with a standard deviation of around 10 frames when we use the ground truth focal length. When we use our predicted focal length, we get an error of up to about 11 frames and a standard deviation of 15 frames for most of the experiments. We report our results in Table \ref{tab:time_sync}.
We discuss one failure case happening for large shifts in the limitations section. 

For the multi-person case, we perform a similar experiment as with the single-person case, except we shift the sequence with offsets 0, 25, and 50 on the EPFL Terrace and Laboratory sequences using our predicted focal lengths. We report our results in Tables \ref{tab:multi_results} and \ref{tab:pair_results}. For these sequences obtain an error within 5 frames of the ground truth offset.


\begin{table}
\centering
\resizebox{1\linewidth}{!}{%
\begin{tabular}{|c|c|c|c|c|c|c|c|}
\cline{2-8}
\multicolumn{1}{c|}{} & \multicolumn{3}{c|}{\textbf{Pred offset (gt f)}} & \multicolumn{3}{c|}{\textbf{Pred offset (Pred f)}} & \multicolumn{1}{c|}{\textbf{\cite{s21072464}*}} \\
\hline
offset gt & mean & median & std & mean & median & std & mean\\ \hline
0        & 4.0 & 1.0 & 10.0
         & 2.90 & 2.5 & 14.05 & 0.0
             \\ \hline
50        & 53.93 & 51 & 10.05
          & 52.81 & 52.5 & 14.12 & N/A
            \\ \hline
100       & 103.88 & 101 & 10.17
          & 102.67 & 102 & 14.22 & N/A
          \\ \hline
150       & 154.5 & 154 & 10.15
          & 153.48 & 153 & 15.04 & N/A
          \\ \hline
200       & 204.76 & 205 & 10.082
          & 189.26 & 201 & 51.70 & N/A
            \\ \hline
\end{tabular}%
}
\caption{\textbf{Temporal synchronization experiments.} This table shows the predicted offset given pairs of videos of Human3.6M with varying offsets as well as their standard deviations. (gt f) means that we run our synchronization using ground truth focal length, and (pred f) means we run our synchronization using predicted focal length. $^*$using GT calibration and not tested on large offsets.}
\label{tab:time_sync}
\end{table}

\subsection{Synchronized Bundle Adjustment Experiments}
To compare against existing methods requiring synchronized cameras, we test on the EPFL Terrace sequence without introducing temporal shift. We compare against Xu et al. \cite{xu2021widebaseline} as well as the other methods that Xu et al. tested including SIFT \cite{10.1023/B:VISI.0000029664.99615.94}, BFM \cite{inproceedings}, SuperPoint \cite{detone2018superpoint}, SuperGlue \cite{sarlin20superglue}, and WxBS \cite{mishkin2015wxbs}. The oracle method in Table \ref{tab:terrace_bundle} simply means that they used manual pose annotations. 
Some methods, such as Sift + BFM, use the first method to extract the keypoints, and the second method to match the keypoints.
The Sift + BFM, Superpoint + BFM, and Oracle methods use Xu et al. Geosolver after finding the correspondences. We find that our method gives reasonable reconstructions compared to the other methods, only being outperformed by Xu et al. and the oracle. However, only by a small margin and they both utilize ground truth intrinsics while we use our estimated intrinsic. 

\begin{table}[]
\centering
\resizebox{0.65\linewidth}{!}{%
\begin{tabular}{|l|c|c|}
\hline
Method                  & mm $\downarrow$ & degree $\downarrow$                 \\ \hline
SIFT \cite{10.1023/B:VISI.0000029664.99615.94} + BFM \cite{inproceedings}             & 4599 & 55.03          \\ \hline
SuperPoint \cite{detone2018superpoint} + \cite{inproceedings}       & 358 & 54.68          \\ \hline
WxBS \cite{mishkin2015wxbs}                           & 1302 & 54.14        \\ \hline
\cite{detone2018superpoint} + SuperGlue \cite{sarlin20superglue} & 9934 & 36.96        \\ \hline
Oracle(Manual-pts)                       & 390& 1.18            \\ \hline
\cite{xu2021widebaseline} (Manual-bbox)                   & 308 & 0.52            \\ \hline
\cite{xu2021widebaseline} (ReID-bbox)                         & 308 & 0.52           \\ \hline
Ours                                     &138 & 1.82                                    \\ \hline
\end{tabular}%
}
\caption{\textbf{Synchronized bundle adjustment experiments.} Camera pose error for the Terrace sequence.}
\label{tab:terrace_bundle}
\end{table}

\subsection{Multiview Offset Experiments}
For multiview offset experiments, we test our method using a similar setting as in the temporal experiments, except we also apply our multiview calibration algorithm afterward in order to analyze the effects of synchronization accuracy on the complete calibration pipeline. As a baseline, we also run our multiview calibration method on the unsynchronized sequences. For these experiments, we do not run our bundle adjustment algorithm since two views provide insufficient constraints under noisy keypoint estimates. For the single-person case, we test it on Human3.6M with 4 cameras for subjects 1,5,6,7,8,9, and 11. We report our results in Table \ref{tab:offset_calibration_table}. We show that although our synchronization method is not perfect, it performs much better than running the multiview calibration algorithm without temporal synchronization. 

For the multi-person case, we utilize the EPFL sequences Terrace and Laboratory which contain up to 4 and 7 people respectively, and 4 cameras each. We proceed with our experiments in a similar manner to the experiments using pairwise cameras, however, we use timesteps 0, 25, and 50. We report our results in Table \ref{tab:multi_results} and Table \ref{tab:pair_results}. Like in the single-person case, we show that our synchronization method results in better performance than running the multiview calibration algorithm without temporal synchronization. 

\begin{table}[]
\centering
\begin{tabular}{|lc|c|c|c|c|}
\hline
\multicolumn{2}{|c|}{offset (gt)} & 0 & 100 & 150 & 200 \\ \hline
\multicolumn{2}{|c|}{offset (pred)} & 5.36 & 105.21 & 156.21 & 191.96 \\ \hline
\multicolumn{1}{|l|}{\multirow{3}{*}{No Sync}} & ° & 5.78 & 36.24 & 42.71 & 73.04 \\ \cline{2-6} 
\multicolumn{1}{|l|}{} & m & 0.057 & 0.26 & 0.46 & 0.82 \\ \cline{2-6} 
\multicolumn{1}{|l|}{} & \%f & 15.06 & 15.06 & 15.06 & 15.06 \\ \hline
\multicolumn{1}{|l|}{\multirow{3}{*}{Sync}} & ° & 10.67 & 10.58 & 10.99 & 16.89 \\ \cline{2-6} 
\multicolumn{1}{|l|}{} & m & 0.066 & 0.065 & 0.068 & 0.21 \\ \cline{2-6} 
\multicolumn{1}{|l|}{} & \%f & 15.06 & 15.06 & 15.06 & 15.06 \\ \hline
\end{tabular}
\caption{\textbf{Multiview offset experiments for Human3.6M.} We report the Camera pose error for unsynchronized sequences and synchronized. No sync means that the algorithm is run without running our synchronization step. Sync means we run our synchronization step. Offset pred means the predicted offset from our synchronization method.}
\label{tab:offset_calibration_table}
\end{table}

\begin{table}[]
\centering
\begin{tabular}{|lc|c|c|c|}
\hline
\multicolumn{2}{|c|}{offset (gt)} & 0 & 25 & 50 \\ \hline
\multicolumn{2}{|c|}{offset (pred)} & 4 & 29 & 55 \\ \hline
\multicolumn{1}{|l|}{\multirow{3}{*}{No Sync}} & ° & 2.73 & 6.32 & 13.07 \\ \cline{2-5} 
\multicolumn{1}{|l|}{} & m & 0.073 & 0.097 & 0.15 \\ \cline{2-5} 
\multicolumn{1}{|l|}{} & \%f & 9.45 & 9.45 & 9.45 \\ \hline
\multicolumn{1}{|l|}{\multirow{3}{*}{Sync}} & ° & 2.14 & 2.15 & 2.18 \\ \cline{2-5} 
\multicolumn{1}{|l|}{} & m & 0.070 & 0.069 & 0.069 \\ \cline{2-5} 
\multicolumn{1}{|l|}{} & \%f & 9.45 & 9.45 & 9.45 \\ \hline
\end{tabular}
\caption{\textbf{Multiview offset experiments for Terrace.} Showing that the synchronization significantly improves camera position and angle. Focal length can be estimated from a single view and is not further refined in this experiment.}
\label{tab:multi_results}
\end{table}

\begin{table}[]
\centering
\begin{tabular}{|lc|c|c|c|}
\hline
\multicolumn{2}{|c|}{offset (gt)} & 0 & 25 & 50 \\ \hline
\multicolumn{2}{|c|}{offset (pred)} & 1.67 & 29 & 55.33 \\ \hline
\multicolumn{1}{|l|}{\multirow{3}{*}{No Sync}} & ° & 0.81 & 10.02 & 44.10 \\ \cline{2-5} 
\multicolumn{1}{|l|}{} & m & 1.15 & 1.16 & 1.15 \\ \cline{2-5} 
\multicolumn{1}{|l|}{} & \%f & 15.63 & 15.63 & 15.63 \\ \hline
\multicolumn{1}{|l|}{\multirow{3}{*}{Sync}} & ° & 0.51 & 0.61 & 0.81 \\ \cline{2-5} 
\multicolumn{1}{|l|}{} & m & 1.15 & 1.15 & 1.15 \\ \cline{2-5} 
\multicolumn{1}{|l|}{} & \%f & 15.63 & 15.63 & 15.63 \\ \hline
\end{tabular}
\caption{\textbf{Multiview offset experiments for the Laboratory sequence.} Improvements on this indoor sequence are consistent with the outdoor terrace sequence in Tab.~\ref{tab:multi_results}.}
\label{tab:pair_results}
\end{table}

\section{Limitations and Discussion}

From the results of our experiments, we demonstrate that our cascaded calibration approach can produce comparable results to other methods while requiring fewer inputs, only requiring a 2D detector instead of having to use ground truth intrinsics, extrinsics, or temporal synchronization, or additional tools such as tracking or reidentification. The results can serve as a starting point for further research since we also publish our code and hyper-parameters such that direct comparisons can be made.

Although we show that our method works well on most scenes, in certain extreme configurations, some of the steps fail and subsequent steps cannot recover due to the assumption that the initialization from the previous step is within error bounds. In future work, we plan to detect such errors and to deploy learning-based solutions to overcome them.
\subsection{Failure Case: Periodic Motion}
In Human36m walking sequences, the people are walking in a circle. This causes the distance curves for 2 views to have a periodic shape which means that for every n frame, the curve repeats itself. This can be problematic for the time synchronization since this would mean multiple offsets can give a similar result. 

\begin{figure}[]
        \centering
    \begin{subfigure}[t]{0.8\linewidth} 
        \includegraphics[width=\linewidth, trim=0.1cm 0 0.01cm 0, clip]{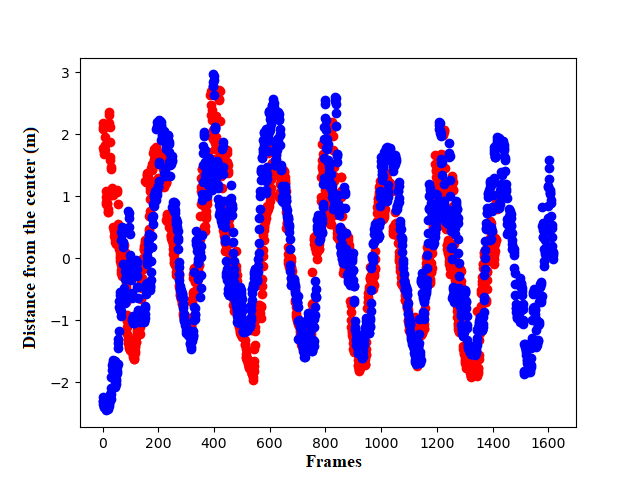}
        \label{fig:failure0}
        \caption{Sequences}
    \end{subfigure}
    \hfill 
    \begin{subfigure}[t]{0.8\linewidth}
        \includegraphics[width=\linewidth, trim=0.1cm 0 0.01cm 0, clip]{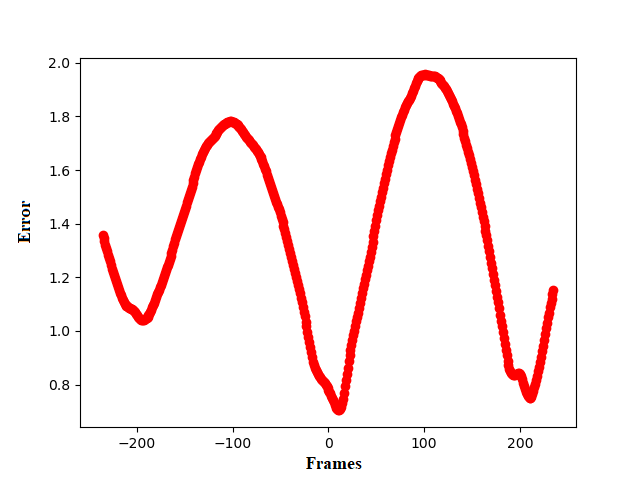}
        \label{fig:failure1}
          \caption{Error curve}
    \end{subfigure}
    \caption{\textbf{Periodic motion.} Due to the period of 200 frames, there are two similar local minimums at 0 and 200.}
    \label{fig:failure}
\end{figure}

Our algorithm's time synchronization module fails on Subject 11 when we set the offset to 200 because in Figure \ref{fig:failure}, we note that the distance curves have a period of about 200 frames, which results in our error curve having two very similar local minimums at around 0 and 200. However, such a large misalignment paired with harmonic repetition is unusual in practice. In future work, the local motion, such as the articulation of arms could be used to further disambiguate frames. However, this is non-trivial as occlusions and different viewing angles have to be considered. 

\section{Conclusion}
\raggedbottom
We designed, implemented, and open-sourced a method to calibrate multiple cameras, even when videos are out of sync. It enables new application domains for 3D vision, where reconstructions require the precision of multi-view but the ease of recording with consumer-grade cameras without hardware synchronization capabilities. Our future works will focus on improving the bundle adjustment step for sparse views in addition to substituting some of our optimization-based steps with learning-based solutions.

\section*{$\phantom{space}$} 
{
}
{
    \small
    \bibliographystyle{ieee}
    \bibliography{main}
}

\newpage
{\centering\Large Supplemental \par}


\section{Camera Calibration Derivation - Details}

We follow the well-established direct linear transform (DLT)~\cite{sutherland1974three} method to solve projective relations. We first write our constraints as a linear system of equations that is solved using Singular Value Decomposition (SVD), up to the unknown scale factor arising from the projection. To reach the form $\mM \vx = 0$, we take the cross product of
\begin{equation}
\vp^\text{img} = \mK \vp^\text{cam},
\text{ where }
\mK = 
\begin{pmatrix}
f & 0 & o_1 \\
0 & f & o_2 \\
0 & 0 & 1
\end{pmatrix}
\label{eq:projection}
\end{equation}
and
\begin{equation}
\vp^{\text{cam}}_{\text{shoulder}} = \vp_\text{ankle} + h\vn 
\end{equation}

and subtract the 2 results to derive
\begin{equation}
\vp_\text{shoulder} \times \mK (\vp_\text{ankle} + \vn \cdot h) - \vp_\text{ankle} \times \mK (\vp_\text{ankle}) = 0
\label{eq:cross_prod}
\end{equation}
with $h$ the person height, $\vn$ the normal direction and $\vp_\text{shoulder}$ and $\vp_\text{ankle}$ the shoulder and ankle positions. In the following we subscript variables with an $x,y,z$ to indicate the x,y,z-coordinates and with a number $1,2,3$ to refer to different person locations.

In matrix form, using $\Delta\vp_x = \vp_x^\text{shoulder} - \vp_x^\text{ankle}$, $\Delta\vp_y = \vp_y^\text{shoulder} - \vp_y^\text{ankle}$, and $z$ to represent the unknown depth of the ankle, Eq.~\ref{eq:cross_prod} can be expressed as

\begin{equation}
\begin{pmatrix}
0 & -1 & \vp_y^\text{shoulder} & 0 & -1 & \Delta\vp_y \\ 1 & 0 & -\vp_x^\text{shoulder} & 1 & 0 & -\Delta\vp_x \\
\end{pmatrix}
\begin{pmatrix}
f\vn_x \\ f\vn_y \\ \vn_z \\ \vn_z\vo_x \\ \vn_z\vo_y \\ z/h
\end{pmatrix}
= 0,
\label{cross_matrix}
\end{equation}
where $f$ is the focal length and $\vo$ the principal point of the camera intrinsics $\mK$.
 By using at least three 2D shoulder $\vp_\text{shoulder}$ and ankle $\vp_\text{ankle}$ detections, we form the constraint matrix
\begin{equation}
\textbf{D} =
\begin{pmatrix}
0 & -1 & \vp_{y1}^\text{shoulder} & \Delta\vp_{y1} & 0 & 0\\ 1 & 0 & -\vp_{x1}^\text{shoulder} & -\Delta\vp_{x1} & 0 & 0\\
0 & -1 & \vp_{y2}^\text{shoulder} & 0 & \Delta\vp_{y2} & 0\\ 1 & 0 & -\vp_{x2}^\text{shoulder} & 0 & -\Delta\vp_{x2} & 0\\
0 & -1 & \vp_{y3}^\text{shoulder} & 0 & 0 & \Delta\vp_{y3}\\ 1 & 0 & -\vp_{x3}^\text{shoulder} & 0 & 0 & -\Delta\vp_{x3}
\end{pmatrix}
\label{eq: full_matrix}
\end{equation}
that gives the system of equations
\begin{equation}
\textbf{D} 
\begin{pmatrix}
f\vn_x + \vn_z\vo_x\\ f\vn_y + \vn_z\vo_y\\ \vn_z \\ z_1/h \\z_2/h \\ z_3/h
\label{eq: DLT}
\end{pmatrix}
= 0
.
\end{equation}

We solve Eq.~\ref{eq: DLT} using SVD. Having more than three ankles and shoulders results in an over-determined system, for which we can find a least-squares solution.

\paragraph{Ground normal extraction.}

Since Eq.~\ref{eq: DLT} is a $6 \times 6$ system with rank five, any solution we find is unique up to a scalar. In order to determine $\vn$ from the SVD or least-squares solution, we use the fact that the normal vector is perpendicular to any vector formed by a pair of ankles. Using

\begin{equation}
    \begin{bmatrix}
    \bar{\vn_x}\\\bar{\vn_y}\\\bar{\vn_z}\\\bar{z_1}\\\bar{z_2}\\\bar{z_3}
    \end{bmatrix}
    = \lambda
    \begin{bmatrix}
    \vn_x + \vn_z\vo_x/f\\\vn_y + \vn_z\vo_y/f\\\vn_z/f\\z_1/(hf)\\z_2/(hf)\\z_3/(hf)
    \end{bmatrix},
    \label{svd_soln_apndx}
\end{equation}
If we do not have a given focal length, we can derive the equation for the focal length,
\begin{equation}
f = \sqrt{\frac{(-(\bar{\vn}_x - \bar{\vn}_z\vo_x)\bar{\vp}_\text{x} - 
(\bar{\vn}_y - \bar{n}_z\vo_y)\bar{\vp}_\text{y})}{(\bar{\vn}_z(\bar{z}_1 - \bar{z}_2))}},
\end{equation}
with 
\begin{equation}
\bar{\vp}_\text{x} = ((\vp_{x1}^\text{ankle} - \vo_x)\bar{z}_1 - (\vp_{x2}^\text{ankle} - \vo_x)\bar{z}_2)
p\end{equation}
and
\begin{equation}
\bar{\vp}_\text{y} = ((\vp_{y1}^\text{ankle} - \vo_y)\bar{z}_1 - (\vp_{y2}^\text{ankle} - \vo_y)\bar{z}_2).
\end{equation}

Either the estimated focal lengths or a given focal length enables us to recover $\lambda\vn$ and $\lambda$($z_1$, $z_2$, $z_3$). To
remove $\lambda$, we divide both vectors by the $L_2$ norm of $\lambda\vn$, giving us a unique $\vn$ of length one and ankle depths $z_1$, $z_2$, and $z_3$.

Using the normal vector $\vn$ and the known depths $z_1$, $z_2$, and $z_3$, we recover the orientation and position of the ground plane.

\section{Filtering}
\label{sec:supp_equations}
Since we require poses that are standing straight up for our single view calibration method, we use \ref{eq:supp_angle_filter} to determine if a pose is standing straight up by measuring the angle of the knee keypoint and the angle of the hip keypoint.
\begin{equation}
\begin{split}
& \hat{\vp}_{\text{right}_{1}} = \vp_{\text{right ankle}}^\text{img} - \vp_{\text{right knee}}^\text{img} \\
& \hat{\vp}_{\text{right}_{2}} = \vp_{\text{right hip}}^\text{img} - \vp_{\text{right knee}}^\text{img} \\
& \hat{\vp}_{\text{right}_{3}} = \vp_{\text{right shoulder}}^\text{img} - \vp_{\text{right knee}}^\text{img} \\
& \hat{\vp}_{\text{left}_{1}} = \vp_{\text{left ankle}}^\text{img} - \vp_{\text{left knee}}^\text{img} \\
& \hat{\vp}_{\text{left}_{2}} = \vp_{\text{left hip}}^\text{img} - \vp_{\text{left knee}}^\text{img} \\
& \hat{\vp}_{\text{left}_{3}} = \vp_{\text{left shoulder}}^\text{img} - \vp_{\text{left knee}}^\text{img} \\
& L_{\text{right}} = |\frac{\hat{\vp}_{\text{right}_{1}} \cdot  \hat{\vp}_{\text{right}_{2}}}{\|\hat{\vp}_{\text{right}_{1}} \|\|\hat{\vp}_{\text{right}_{2}}\|} - \pi| + |\frac{\hat{\vp}_{\text{right}_{2}} \cdot \hat{\vp}_{\text{right}_{3}}}{\|\hat{\vp}_{\text{right}_{2}} \|\|\hat{\vp}_{\text{right}_{3}}\|} - \pi| \\
& L_{\text{left}} = |\frac{\hat{\vp}_{\text{left}_{1}} \cdot  \hat{\vp}_{\text{left}_{2}}}{\|\hat{\vp}_{\text{left}_{1}} \|\|\hat{\vp}_{\text{left}_{2}}\|} - \pi| + |\frac{\hat{\vp}_{\text{left}_{2}} \cdot  \hat{\vp}_{\text{left}_{3}}}{\|\hat{\vp}_{\text{left}_{2}} \|\|\hat{\vp}_{\text{left}_{3}}\|} - \pi| \\
& \text{Filter}(p^{\text{img}}) = \text{min}(L_{\text{right}}, L_{\text{left}})
\end{split}
\label{eq:supp_angle_filter}
\end{equation}

\section{2D Rotation Search - Details}
We compute our error using Equation \ref{plane error}, where we augment our detections $\vp^{\text{plane}}_i$ with the time step, which we notate as $\hat{\vp}^{\text{plane}}_i = (x, y, t)$ we define $c_1$ and $c_2$ as camera 1 and camera 2 or the ref camera and the sync camera. We also define $\hat{F}$, $\hat{P}_{c, j}$, and $\hat{K}_{c, j, p}$ to represent the sets of indices of the frames, poses for view $c$ and frame $i$, and keypoints for view $c$, frame $i$, and pose $p$.

\begin{equation}
\begin{split}
& L_{\text{min}}(\text{c}_1, \text{c}_2, i, k_1) = \min_{k_2 \in \hat{P}_{c_2, i}} \| \hat{\vp}^{\text{plane}}_{\text{c}_1, i, k_1, \text{ankle}}-\hat{\vp}^{\text{plane}}_{\text{c}_2, i, k_2,  \text{ankle}} \| \\
& L_{\text{t}}(\text{c}_1, \text{c}_2) = \sum_{i \in \hat{F},k_1 \in \hat{P}_{c_1, i}} L_{\text{min}}(\text{c}_1, \text{c}_2, i, k_1) \\
& L_{\text{rot}}(\text{c}_1, \text{c}_2) = L_{\text{t}}(\text{c}_1, \text{c}_2) + L_{\text{t}}(\text{c}_2, \text{c}_1)
\end{split}
\label{plane error}
\end{equation}

\section{Bundle Adjustment - Details}
The terms in our bundle adjustment equation are the intersection error $L_{\text{3D}}$, left and right joint symmetry error $ L_{\text{left,right}}$, the height error $L_h$, and the constraint that the ankle is on the plane $L_p$. The intersection error \ref{eq:intersection_error} is simply the distance of the closest points from the reprojection rays. The left and right joint symmetry error \ref{eq:left_right_error} constrains the right and left joints to have the same length. We also experimented with the height error \ref{eq:h_error} that constrains the joints connecting from the ankle to the neck, to sum up to the assumed height $h$ and the \ref{eq:foot_error} constrain the foot to be on the plane by making the z component of the ankle coordinates 0. 

\begin{equation}
L_{\text{3D}}^{\text{c}_1, \text{c}_2}  = \frac{\sum_{j \in \hat{F},p \in \hat{P}_{j},k \in \hat{K}_{j, p}} \lVert \vp^{\text{world}}_{\text{c}_1,j,p,k} - \vp^{\text{world}}_{\text{c}_2,j,p,k} \rVert_2}{|\hat{F}||\hat{P}||\hat{K}|}
\label{eq:intersection_error}
\end{equation}

\begin{equation}
L_{\text{left,right}}^{\text{c}_1, \text{c}_2}  = \frac{\sum_{j \in \hat{F},p \in \hat{P}_{j},k \in \hat{K}_{j, p}} | \lVert J^{\text{world}}_{\text{c}_1,j,p,k}\rVert_2 - \lVert J^{\text{world}}_{\text{c}_2,j,p,k} \rVert_2 |}  {|\hat{F}||\hat{P}||\hat{K}|}
\label{eq:left_right_error}
\end{equation}

\begin{equation}
L_{\text{h}}^{\text{c}}  = \frac{\sum_{j \in \hat{F},p \in \hat{P}_{j},k \in \hat{K}_{j, p}} \lVert J^{\text{world}}_{\text{c}_1,j,p,k}\rVert_2}  {|\hat{F}||\hat{P}||\hat{K}|}
\label{eq:h_error}
\end{equation}

\begin{equation}
L_{\text{p}}^{\text{cam}}  = \frac{\sum_{j \in \hat{F},p \in \hat{P}_{j}} \lVert \vp^{\text{world}}(z)_{\text{c},j,p,\text{ankle}} \rVert}{|\hat{F}||\hat{P}||\hat{K}|}
\label{eq:foot_error}
\end{equation}

\section{Experiments}
\subsection{Hyperparameters}
We use the EPFL terrace2 as a validation set for our hyperparameters.
\label{sec:supp_experiments}
\subsection{Terrace Calibration}
\label{sec:terrace_calibration}
Although the terrace sequence cameras are calibrated for the terrace sequence, the dataset has two versions of the calibrations. In one of them, instead of the usual intrinsic and extrinsic matrices, the calibration files contain 2 homographies representing the transformation to head plane and ankle plane. The other one contains intrinsic extrinsic calibrations using the Tsai calibration method \cite{tsai1987}, however, the calibrations are based on a different-sized image than the ones in the dataset. To rectify this, we use the homographies to create a virtual checkboard on the ground plane as well as the image dimension in the data and use opencv to compute the camera matrix. 

\subsection{Temporal Synchronization Experiments}
In this section, we show our results for temporal synchronization experiments on Human3.6M for each subject. We report our results in Table \ref{tab:supp_time_sync}.

\begin{table}[]
\begin{tabular}{|l|l|l|l|}
\hline
Subject & Shift (gt) & \begin{tabular}[c]{@{}l@{}}Shift \\ (gt focal)\end{tabular} & \begin{tabular}[c]{@{}l@{}}Shift \\ (pred focal)\end{tabular} \\ \hline
S1      & 0        & 11.24         & 11.82           \\ \hline
S1      & 50       & 54.10         & 54.25              \\ \hline
S1      & 100      & 104.10         & 104.25             \\ \hline
S1      & 150      & 153.72         & 154.28           \\ \hline
S1      & 200      & 204.18         & 204.28           \\ \hline
S5      & 0        & 12.84         & 14.997           \\ \hline
S5      & 50       & 55.49         & 55.68             \\ \hline
S5      & 100      & 105.38        & 105.48           \\ \hline
S5      & 150      & 155.38         & 155.28            \\ \hline
S5      & 200      & 205.25         & 204.97           \\ \hline
S6      & 0        & 13.94         & 21.75           \\ \hline
S6      & 50       & 56.08         & 60.03             \\ \hline
S6      & 100      & 106.14         & 110.18            \\ \hline
S6      & 150      & 156.41         & 166.31           \\ \hline
S6      & 200      & 206.51         & 215.91           \\ \hline
S7      & 0        & 14.14         & 13.22           \\ \hline
S7      & 50       & 56.77         & 54.34           \\ \hline
S7      & 100      & 106.90         & 104.28           \\ \hline
S7      & 150      & 157.24         & 154.8              \\ \hline
S7      & 200      & 206.93         & 204.90           \\ \hline
S8      & 0        & 15.71         & 14.46           \\ \hline
S8      & 50       & 58.82         & 58.42           \\ \hline
S8      & 100      & 108.58          & 106.79           \\ \hline
S8      & 150      & 158.82         & 158.67           \\ \hline
S8      & 200      & 208.92         & 208.67           \\ \hline
S9      & 0        & 3.37         & 10.73           \\ \hline
S9      & 50       & 52.63         & 52.15           \\ \hline
S9      & 100      & 101.06         & 102.31           \\ \hline
S9      & 150      & 150.39         & 152.98           \\ \hline
S9      & 200      & 201.20         & 204.02           \\ \hline
S11     & 0        & 9.22         & 12.71           \\ \hline
S11     & 50       & 57.17         & 54.16           \\ \hline
S11     & 100      & 107.58          & 104.02           \\ \hline
S11     & 150      & 159.12         & 154.18           \\ \hline
S11     & 200      & 209.28         & 70.31           \\ \hline
\end{tabular}
  \caption{\textbf{Temporal synchronization experiments.} We report our results for each subject for Human3.6M. GT shift represents the ground truth offset, while GT focal represents running our method with ground truth focal length, and Pred focal represents running our method with the focal length predicted by our method. Results are in frame offsets.}
  \label{tab:supp_time_sync}
\end{table}

\section{Simulations}
\label{section:Simulations}

\begin{figure}[t!]
\centering
\includegraphics[width=0.98\linewidth,trim={0.1cm 0 0.01cm 0},clip]{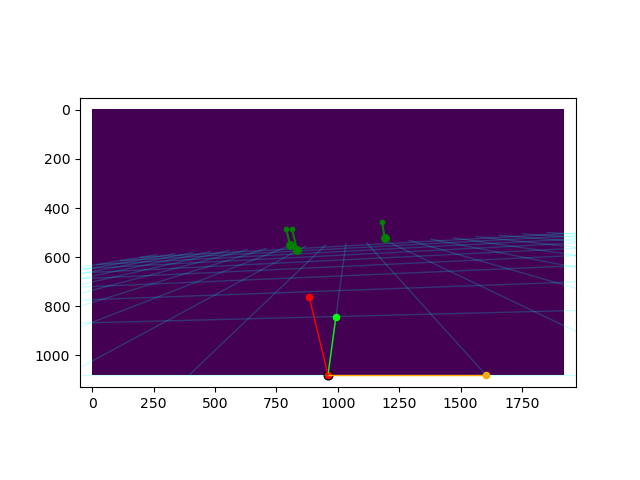}
  \caption{\textbf{Simulation.} This shows an example of a scene generated by our simulation where the green dots in the distance represent ankle and shoulder center detections, and the blue lines represent the ground plane.}
\label{fig:simulation} 
\end{figure}

We conduct simulation experiments the same as \cite{fei2021single}, using the same or similar specifications that were given in their simulation study. In our simulated experiments, we seek to test the effects of measurement noise, height variations, and number of people on our single view calibration algorithm, without the effects of detection noise present from pose detectors. This also allows us to get ground truth heights of the people, which we would not have access to in any of the datasets. We generate a scene, as shown in Fig. \ref{fig:simulation}, with a random ground plane and random shoulder and ankle center detections. We specify image dimensions 1920.0 by 1080.0 with focal length $f_X = 960$ and $f_y = 540$, which corresponds to a 90 degree FOV. We run each experiment for 5000 trials.

\parag{Metrics.} 
\begin{itemize}[noitemsep,nolistsep]
\item $\mathbf{Focal}$ $  \mathbf{error}$: the percent focal error is simply computed by $100 \cdot \frac{ |f_{\text{gt} - f_{\text{pred}}}|}{f_{\text{gt}}}$. This is represented in the tables as $f_x\%$ and $f_y\%$
\item $\mathbf{Normal}$ $  \mathbf{error}$: The degree normal error is the angle difference between the ground truth and predicted normal vectors, represented in the table as $N (°)$.
\item $\mathbf{Position}$ $  \mathbf{error}$: The position error is computed by taking the taking the predicted Euclidean distance from the camera to the ground plane $\rho_\text{pred}$, and computing $\frac{|\rho_{\text{pred}} - \rho_{\text{gt}}|}{\rho_{\text{gt}}}$. This is represented by $\rho\%$.
\item $\mathbf{Reconstuction}$ $  \mathbf{error}$: The reconstruction error is the percent error between Euclidean distance between the predicted 3D points and the ground truth 3D points divided by the ground truth distance to the camera. This is represented in the table by $X\%$
\item $\mathbf{Failure}$ $ \mathbf{rate}$: Since this could lead to an unsolvable linear system, we record the failure rate which is the percentage of noisy systems that are unsolvable. This is represented as $\text{fail}\%$
\end{itemize}

\paragraph{Measurement Noise Trials}
For these experiments, we fix the height to be 1.6 meters and use 3 pairs of shoulder and ankle center positions. We add a zero mean Gaussian with varying standard deviations to these generated positions. We solve our DLT equations with a height of 1.6 meters. We record our results in Table \ref{tab:Measurement noise}. For these trials, adding zero error results in an error for all the metrics that were virtually 0. As the noise standard deviation grows, the error for all metrics increases as expected.

\paragraph{Height Variation Trials}
For these experiments, we fix the detection noise to be sampled from a zero mean Gaussian with a standard deviation of 0.5 and we sample heights from a Gaussian centered at 1.7 meters with varying standard deviations. In our algorithm, we solve the DLT equations with a height equal to 1.7m. We report our results in Table \ref{tab:height experiments}. As expected, as the standard deviations of the heights increase, the worse the results become since our algorithm uses a fixed height of 1.7 meters.

\paragraph{Number of People Trials}
For these experiments, we fix the detection noise as in the height variation trails, but we sample the height from a Gaussian centered at 1.7 meters with a standard deviation of 0.1 meters. Within our DLT equations, we fix the height to be 1.7 meters like the Gaussian mean. We vary the number of pairs of shoulder and ankle center detection and record our results in Table \ref{tab:Number of people}. As the number of people increases, the system becomes more and more over determined which reduces the resulting error.
\begin{table}[htb]
\centering
\begin{tabular}{l|llllll|}
\cline{2-7}
\multicolumn{1}{l|}{}                                   & \multicolumn{6}{l|}{Simulation measurement noise std. in pixels}                                                                                               \\ \hline
\multicolumn{1}{|l|}{Error}                    & \multicolumn{1}{l|}{0.1}  & \multicolumn{1}{l|}{0.2}  & \multicolumn{1}{l|}{0.5}  & \multicolumn{1}{l|}{1.0}   & \multicolumn{1}{l|}{2.0}   & 5.0   \\ \hline
\multicolumn{1}{|l|}{$f_x\%$}       & \multicolumn{1}{l|}{1.85} & \multicolumn{1}{l|}{3.45} & \multicolumn{1}{l|}{6.88} & \multicolumn{1}{l|}{9.70}  & \multicolumn{1}{l|}{16.83} & 27.03 \\  \hline
\multicolumn{1}{|l|}{$f_y\%$}       & \multicolumn{1}{l|}{1.08} & \multicolumn{1}{l|}{1.81} & \multicolumn{1}{l|}{3.66} & \multicolumn{1}{l|}{6.01}  & \multicolumn{1}{l|}{10.13} & 17.75 \\  \hline
\multicolumn{1}{|l|}{{N (°)}}      & \multicolumn{1}{l|}{0.15} & \multicolumn{1}{l|}{0.26} & \multicolumn{1}{l|}{0.60} & \multicolumn{1}{l|}{1.09}  & \multicolumn{1}{l|}{2.22}  & 4.30  \\ \hline
\multicolumn{1}{|l|}{$\rho\%$}            & \multicolumn{1}{l|}{1.58} & \multicolumn{1}{l|}{0.71} & \multicolumn{1}{l|}{1.77} & \multicolumn{1}{l|}{3.23}  & \multicolumn{1}{l|}{7.22}  & 22.52 \\ \hline
\multicolumn{1}{|l|}{$X\%$}          & \multicolumn{1}{l|}{1.06} & \multicolumn{1}{l|}{1.74} & \multicolumn{1}{l|}{3.57} & \multicolumn{1}{l|}{5.92}  & \multicolumn{1}{l|}{10.08} & 17.64 \\ \hline
\multicolumn{1}{|l|}{\text{fail}$\%$}       & \multicolumn{1}{l|}{0.82} & \multicolumn{1}{l|}{1.52} & \multicolumn{1}{l|}{3.4}  & \multicolumn{1}{l|}{5.5}   & \multicolumn{1}{l|}{8.04}  & 17.36 \\ \hline
\end{tabular}
\caption{\textbf{Measurement noise.} We show the error from our calibration for varying measurement noise standard deviations.}
\label{tab:Measurement noise}
\end{table}

\begin{table}[htb]
\centering
\begin{tabular}{l|lllll|}
\cline{2-6}
\multicolumn{1}{l|}{}  & 
\multicolumn{5}{l|}{Number of people} 

\\ \hline
\multicolumn{1}{|l|}{Error}                     & \multicolumn{1}{l|}{5}        & \multicolumn{1}{l|}{10}       & \multicolumn{1}{l|}{20}       & \multicolumn{1}{l|}{50}       & 100      \\ \hline
\multicolumn{1}{|l|}{$f_x\%$}      & \multicolumn{1}{l|}{27.33}    & \multicolumn{1}{l|}{27.68} & \multicolumn{1}{l|}{26.29}    & \multicolumn{1}{l|}{18.21}    & 21.98       \\ \hline
\multicolumn{1}{|l|}{$f_y\%$}     & \multicolumn{1}{l|}{32.40}    & \multicolumn{1}{l|}{25.64}    & \multicolumn{1}{l|}{25.25}    & \multicolumn{1}{l|}{14.94}    & 14.17  \\ \hline
\multicolumn{1}{|l|}{{N (°)}}     & \multicolumn{1}{l|}{3.74}     & \multicolumn{1}{l|}{3.33}     & \multicolumn{1}{l|}{2.62}     & \multicolumn{1}{l|}{1.76}     & 1.13     \\ \hline
\multicolumn{1}{|l|}{$\rho\%$}     & \multicolumn{1}{l|}{21.46} & \multicolumn{1}{l|}{21.66} & \multicolumn{1}{l|}{18.92} & \multicolumn{1}{l|}{21.64} & 24.09 \\ \hline
\multicolumn{1}{|l|}{$X\%$}     & \multicolumn{1}{l|}{20.96} & \multicolumn{1}{l|}{21.24} & \multicolumn{1}{l|}{20.65} & \multicolumn{1}{l|}{30.07} & 42.82 \\ \hline
\multicolumn{1}{|l|}{\text{fail}$\%$}     & \multicolumn{1}{l|}{15.98}    & \multicolumn{1}{l|}{12.2}     & \multicolumn{1}{l|}{8.8}      & \multicolumn{1}{l|}{5.72}     & 3.78     \\ \hline
\end{tabular}
\caption{\textbf{Number of people.} We show the error from our calibration for varying numbers of people}
\label{tab:Number of people}
\end{table}

\begin{table}[htb]
\centering
\begin{tabular}{l|lllll|}
\cline{2-6} & \multicolumn{5}{l|}{Std. of height in meters} \\ \hline
\multicolumn{1}{|l|}{Error}                      & \multicolumn{1}{l|}{0.05}     & \multicolumn{1}{l|}{0.1}      & \multicolumn{1}{l|}{0.15}     & \multicolumn{1}{l|}{0.2}      & 0.25     \\ \hline
\multicolumn{1}{|l|}{$f_x\%$}       & \multicolumn{1}{l|}{7.00} & \multicolumn{1}{l|}{8.75} & \multicolumn{1}{l|}{10.36} & \multicolumn{1}{l|}{10.71} & 13.22   \\ \hline
\multicolumn{1}{|l|}{$f_y\%$}        & \multicolumn{1}{l|}{3.64} & \multicolumn{1}{l|}{4.258} & \multicolumn{1}{l|}{5.27} & \multicolumn{1}{l|}{6.04} & 6.48 \\ \hline
\multicolumn{1}{|l|}{{N (°)}}         & \multicolumn{1}{l|}{0.58} & \multicolumn{1}{l|}{0.63} & \multicolumn{1}{l|}{0.82} & \multicolumn{1}{l|}{0.87} & 1.05 \\ \hline
\multicolumn{1}{|l|}{$\rho\%$}       & \multicolumn{1}{l|}{6.98} & \multicolumn{1}{l|}{10.388} & \multicolumn{1}{l|}{13.29} & \multicolumn{1}{l|}{16.99} & 21.43 \\ \hline
\multicolumn{1}{|l|}{$X\%$}         & \multicolumn{1}{l|}{6.69} & \multicolumn{1}{l|}{10.85} & \multicolumn{1}{l|}{14.60} & \multicolumn{1}{l|}{18.048} & 21.54 \\ \hline
\multicolumn{1}{|l|}{\text{fail}$\%$}     & \multicolumn{1}{l|}{2.62}     & \multicolumn{1}{l|}{3.5}      & \multicolumn{1}{l|}{4.18}     & \multicolumn{1}{l|}{4.7}      & 5.98 \\ \hline
\end{tabular}
\caption{\textbf{Height.} We show the error from our calibration for varying random heights that are drawn from a Gaussian distribution with varying standard deviations.}
\label{tab:height experiments}
\end{table}

\section{Noise propagation from stage to stage}

We conduct experiments to investigate the effects of noise on our pipeline on the EPFL terrace sequence \cite{4359319} on cameras \texttt{terrace1-c0\_avi}, \texttt{terrace1-c1\_avi}, and \texttt{terrace1-c2\_avi}. We run 4 sets of experiments where we take in as input a height parameter and 2D detections, and then we monitor how noise added to different stages of the pipeline propagates to the intermediate and final outputs. In separate experiments, we add noise to the 2D detections, the focal length, the normal vector, and the time synchronization.

\begin{figure*}
  \centering

  \begin{subfigure}{0.32\textwidth}
    \centering
    \includegraphics[width = \textwidth]{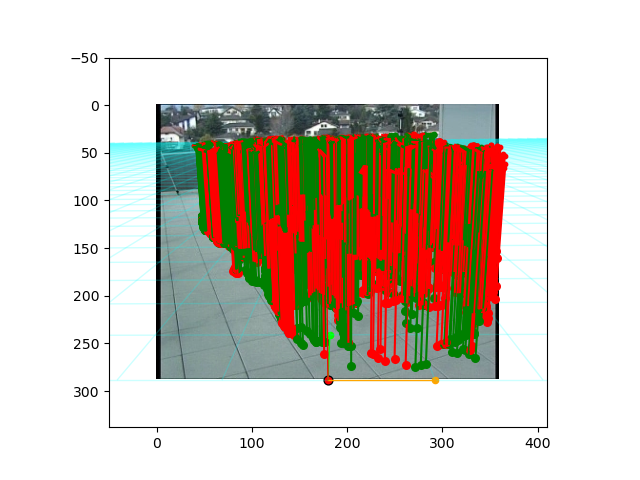}
  \caption{\textbf{No noise}}
\label{fig:noise_0} 
  \end{subfigure}
  \begin{subfigure}{0.32\textwidth}
    \centering
    \includegraphics[width = \textwidth]{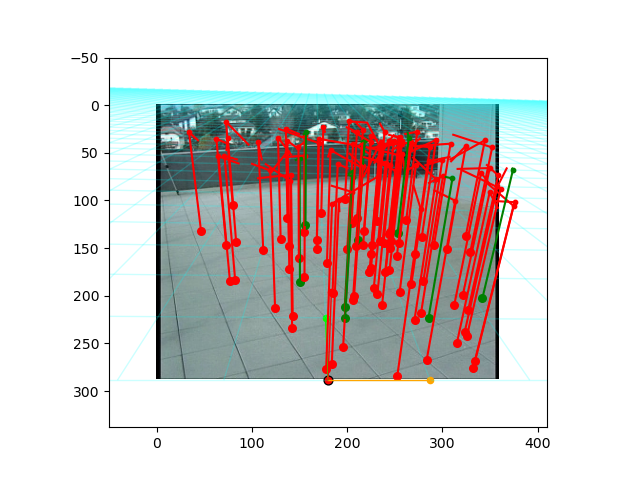}
  \caption{\textbf{Detection noise} $\mu$=0 $\sigma$=15 pixels.}
\label{fig:noise_15} 
  \end{subfigure}
  \begin{subfigure}{0.32\textwidth}
    \centering
    \includegraphics[width =\textwidth]{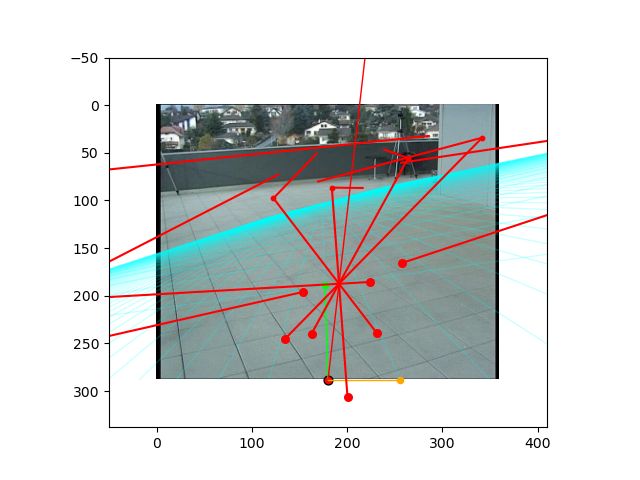}
  \caption{\textbf{Detection noise} $\mu$=0 $\sigma$=35 pixels.}
    \label{fig:noise_35} 
  \end{subfigure}

  \caption{\textbf{Terrace detection noise visualization.} Experiments were conducted on terrace1-c2 with noise sampled from a Gaussian distribution. Here, the green lines represent inlier detections and the red lines represent the outlier detections.}
  \label{fig:Terrace_Detection_Noise_images}
\end{figure*}

\begin{figure}
  \centering
  \hfill
  \begin{subfigure}{0.5\textwidth}
    \centering
\includegraphics[width=0.98\linewidth,trim={0.1cm 0 0.01cm 0},clip]{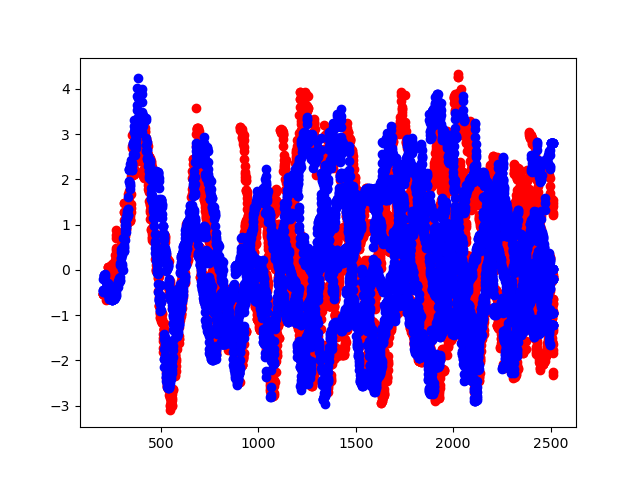}
  \caption{\textbf{Time synchronization} in the 0 detection noise case.}
\label{fig:time_0} 
\label{fig:noise_15} 
  \end{subfigure}

  \vspace{1em} 

  \begin{subfigure}{0.5\textwidth}
    \centering
    \includegraphics[width=0.98\linewidth]{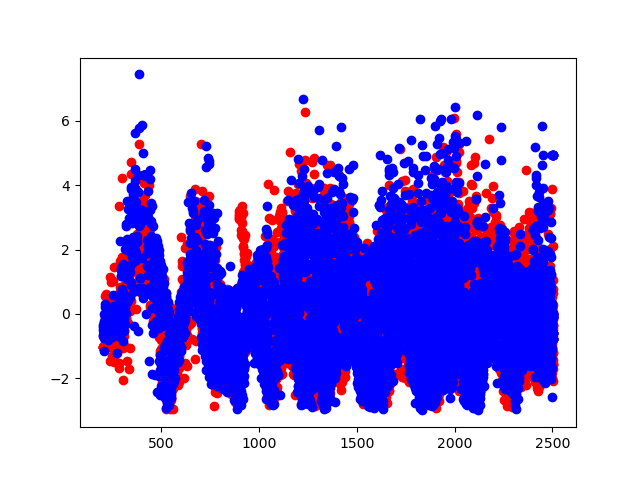}
  \caption{\textbf{Time synchronization} in the 10-pixel std detection noise case. In the zero noise case, we see well-defined curves, however, with noise, outliers cause them to be less well-defined.}
\label{fig:time_25} 
  \end{subfigure}

  \caption{Terrace detection noise level visualization.}
  \label{fig:Terrace_Detection_Noise_time}
\end{figure}

\begin{figure}
  \centering

  \begin{subfigure}{0.5\textwidth}
    \centering
    \includegraphics[width=0.8\linewidth]{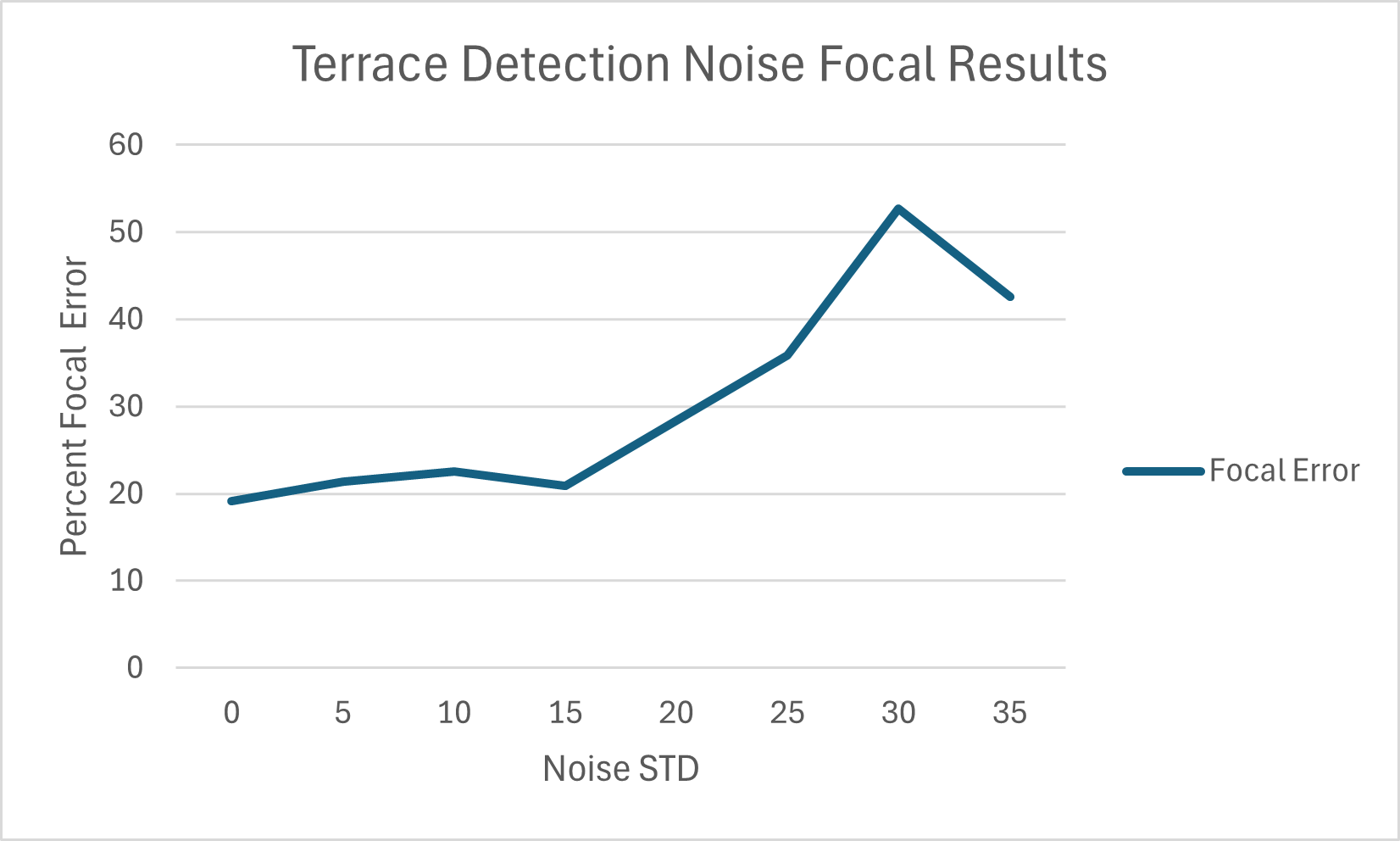}
    \caption{Percent focal length error.}
    \label{fig:Terrace_Detection_Focal_Angle_Results}
  \end{subfigure}
  \hfill
  \begin{subfigure}{0.5\textwidth}
    \centering
    \includegraphics[width=0.8\linewidth]{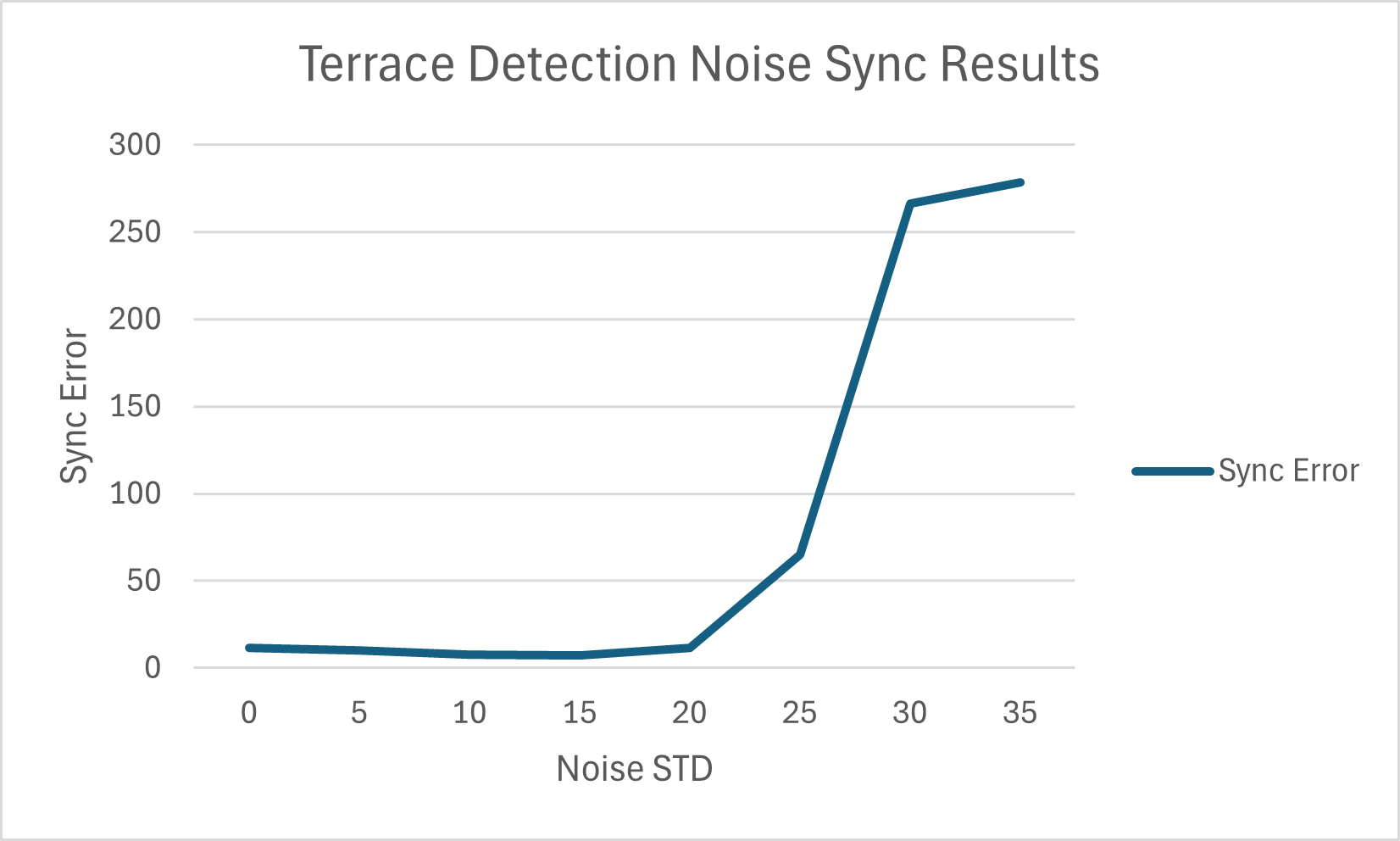}
    \caption{Temporal synchronization error}
    \label{fig:Terrace_Detection_Noise_Sync_Results}
  \end{subfigure}

  \vspace{1em} 

  \begin{subfigure}{0.5\textwidth}
    \centering
    \includegraphics[width=0.8\linewidth]{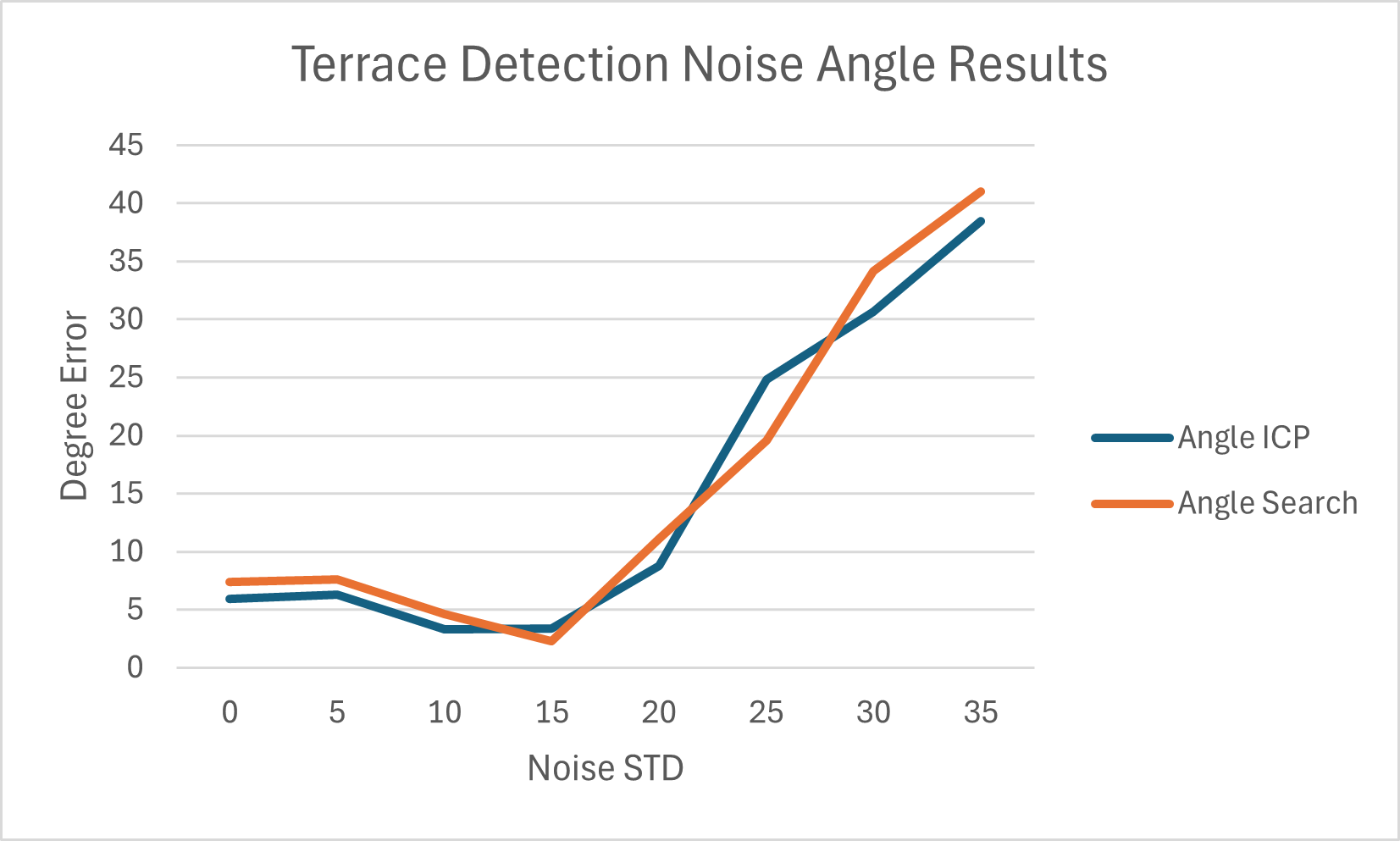}
    \caption{Degree rotation angle error.}
    \label{fig:Terrace_Detection_Noise_Angle_Results}
  \end{subfigure}
  \hfill
  \begin{subfigure}{0.5\textwidth}
    \centering
    \includegraphics[width=0.8\linewidth]{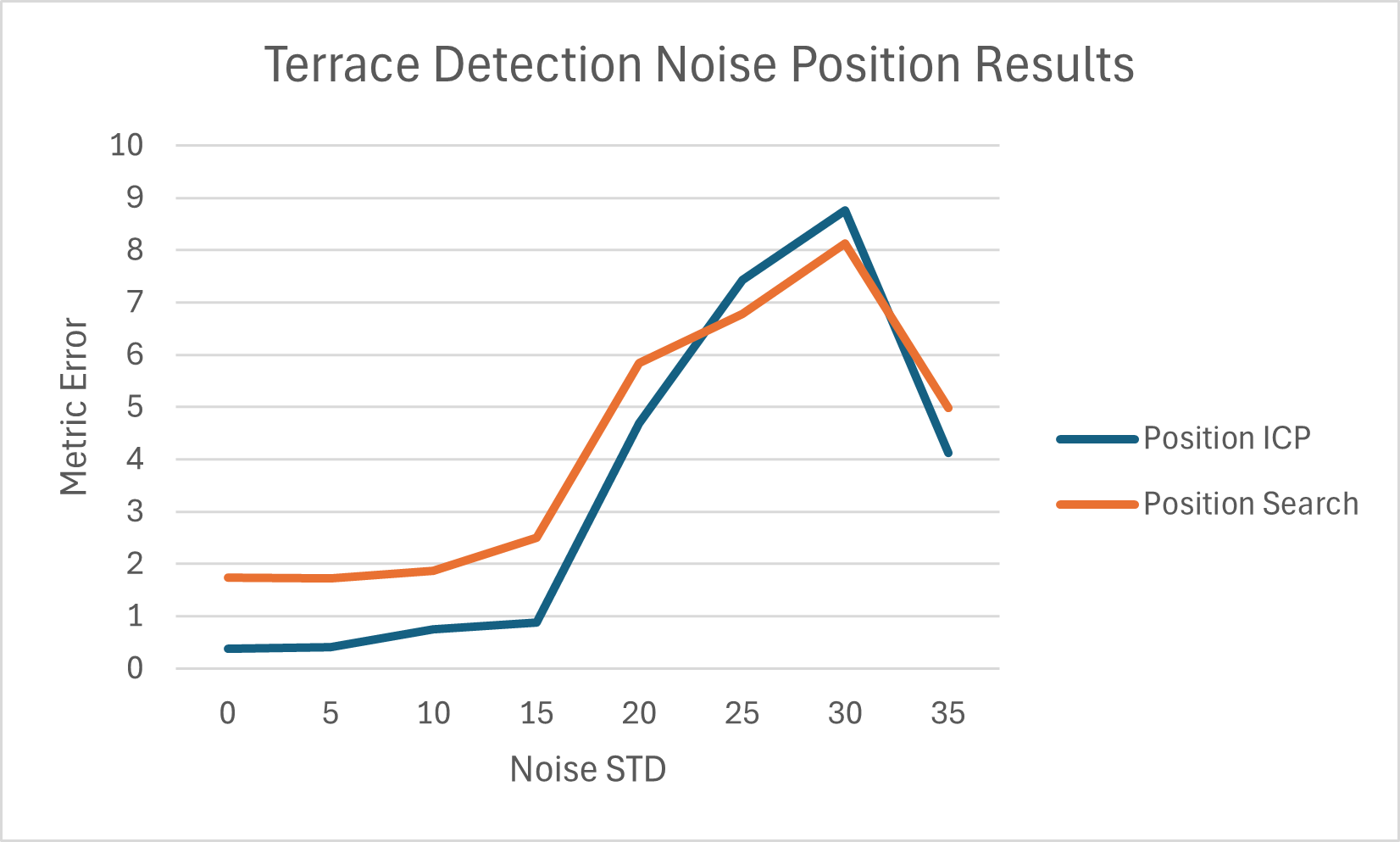}
    \caption{Normalized position error.}
    \label{fig:Terrace_Detection_Noise_Position_Results}
  \end{subfigure}

  \caption{\textbf{Terrace detection noise error curves.} We plot the error on all of our metrics compared to the standard deviation of the noise added to the 2D detections. Note that the errors are non-zero without added noise (std=0) due to the original estimation noise.}
  \label{fig:Terrace_Detection_Noise}
\end{figure}

\paragraph{Detection noise}
For these experiments, we add random noise to the 2D detections sampled from a Gaussian with mean zero and standard deviation 0, 5, 10, 15, 20, 25, 30, and 35 pixels. We run the experiment 5 times for each standard deviation and take the median results. 
We plot the results as error curves in Fig. \ref{fig:Terrace_Detection_Noise}.
What we find is that from 0 to 15 pixels the results for each component are relatively close to each other. However, for standard deviations larger than 15, the error for each component increases significantly since the initial steps of finding the focal length and time synchronization start to fail, resulting in the later stages having unrecoverable errors. We can see that in the zero noise case in Figure \ref{fig:noise_0}, we have a much larger amount of detections that can be used for the single view calibration whereas with the higher noise case, in Figure \ref{fig:noise_35} and \ref{fig:noise_15}, fewer detections are available for calibration because the additional noise causes some poses to appear to be non-standing, which gets filtered out before our RANSAC step. The available ones are mostly outliers in the RANSAC loop. For time synchronization, we can see from Figure \ref{fig:time_0} and Figure \ref{fig:time_25} that as the error standard deviation increases, the time curves become less defined and thus cause the time synchronization step to become less accurate. 

\begin{figure}
  \centering

  \begin{subfigure}{0.5\textwidth}
    \centering
    \includegraphics[width=0.8\linewidth]{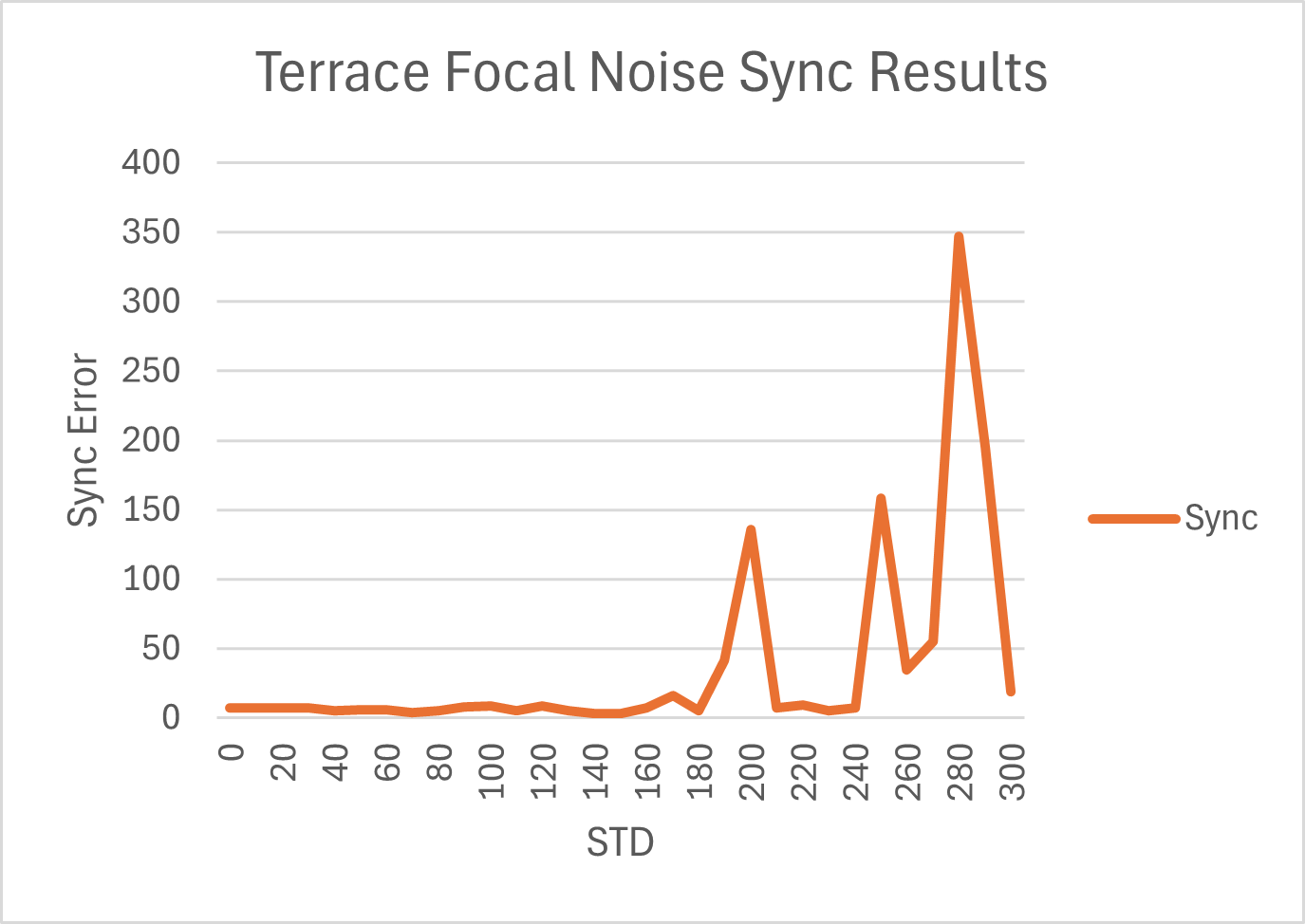}
    \caption{Temporal synchronization error.}
    \label{fig:Terrace_Focal_Noise_Sync_Results}
  \end{subfigure}

  \vspace{1em} 

  \begin{subfigure}{0.5\textwidth}
    \centering
    \includegraphics[width=0.8\linewidth]{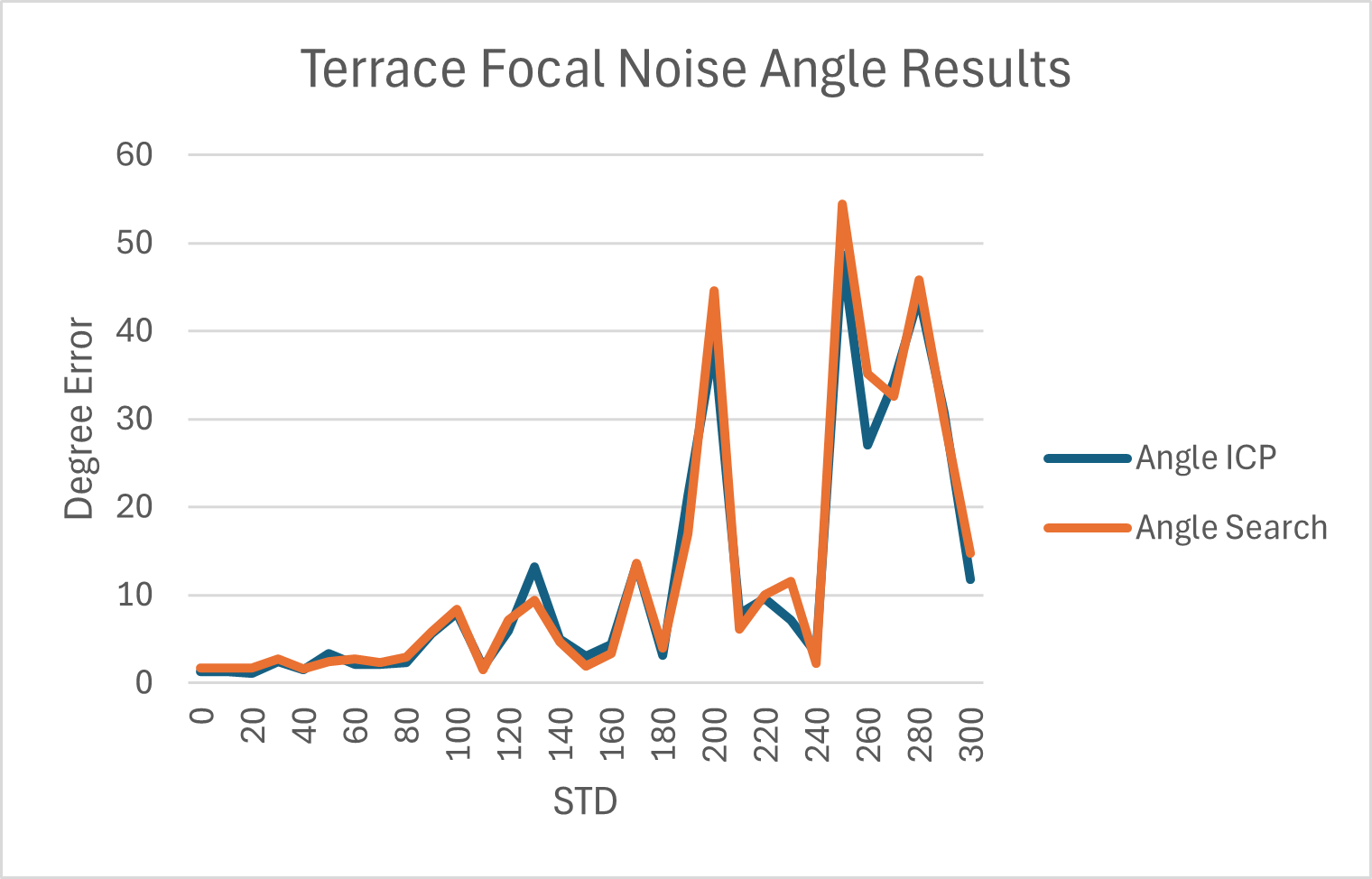}
    \caption{Degree rotation angle error.}
    \label{fig:Terrace_Focal_Noise_Angle_Results}
  \end{subfigure}
  \hfill
  \begin{subfigure}{0.5\textwidth}
    \centering
    \includegraphics[width=0.8\linewidth]{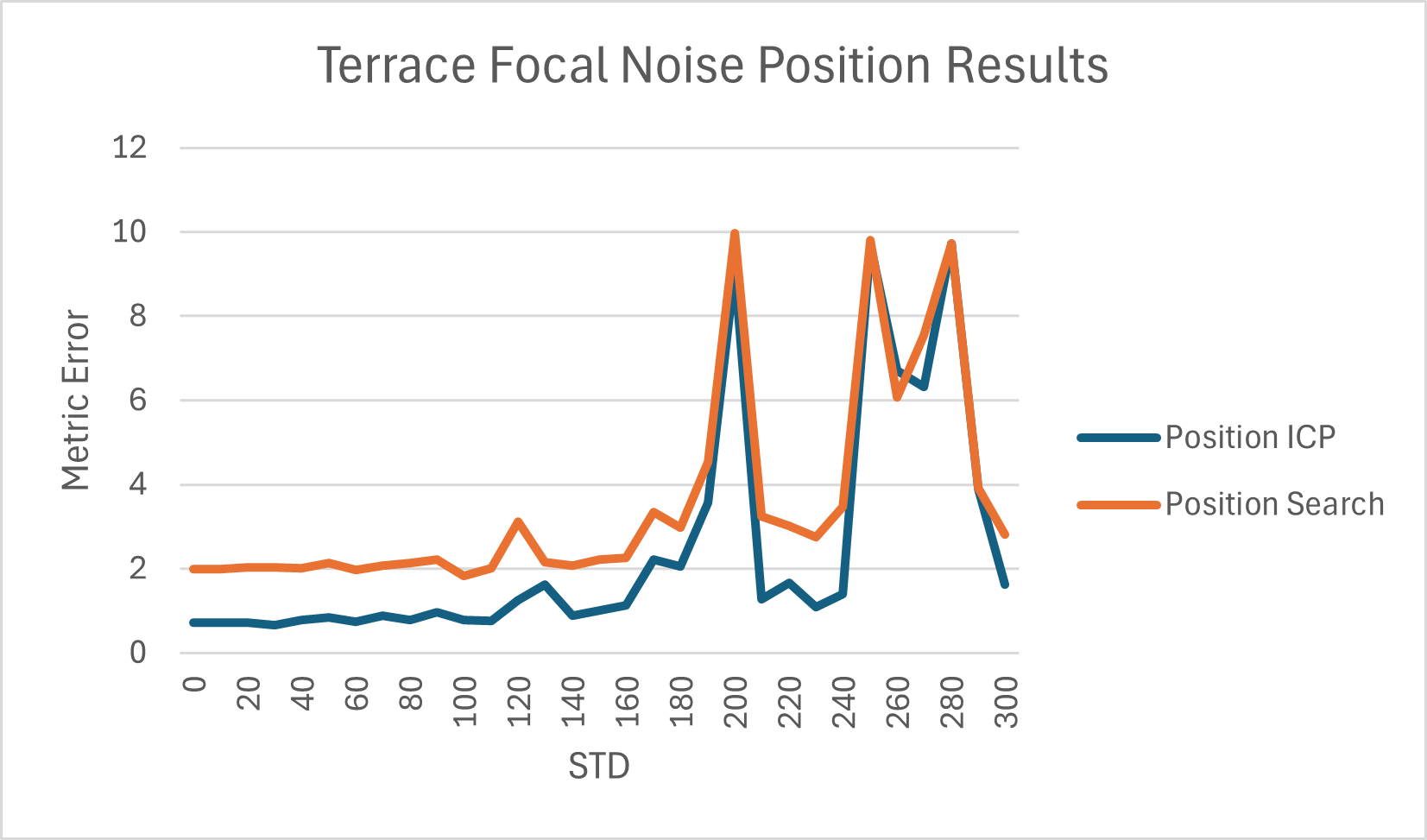}
    \caption{Normalized position error.}
    \label{fig:Terrace_Focal_Noise_Position_Results}
  \end{subfigure}

  \caption{\textbf{Terrace focal noise error curves.} We plot the error on all of our metrics compared to the offset added to the focal length.}
  \label{fig:Terrace_Focal_Noise}
\end{figure}

\paragraph{Focal Length Noise}
For these experiments, we sampled random noise from a Gaussian with mean zero and standard deviation between 0 and 200 pixels to the ground truth focal length while using the ground truth normal vector. We run the experiments five times per standard deviation and take the median result. 
We find that larger magnitudes of offsets result in worse rotation, translation, and temporal synchronization results for the following steps. From Fig. \ref{fig:Terrace_Focal_Noise} we see that the errors do not increase significantly until we use standard deviations beyond 80 pixels. Since the average ground truth focal length is about 437 pixels, we show that even if our predicted focal length is off by as much as 20 percent, our system can still produce reasonable results for the other parameters estimated by the subsequent steps.

\begin{figure}
  \centering

  \begin{subfigure}{0.5\textwidth}
    \centering
    \includegraphics[width=0.8\linewidth]{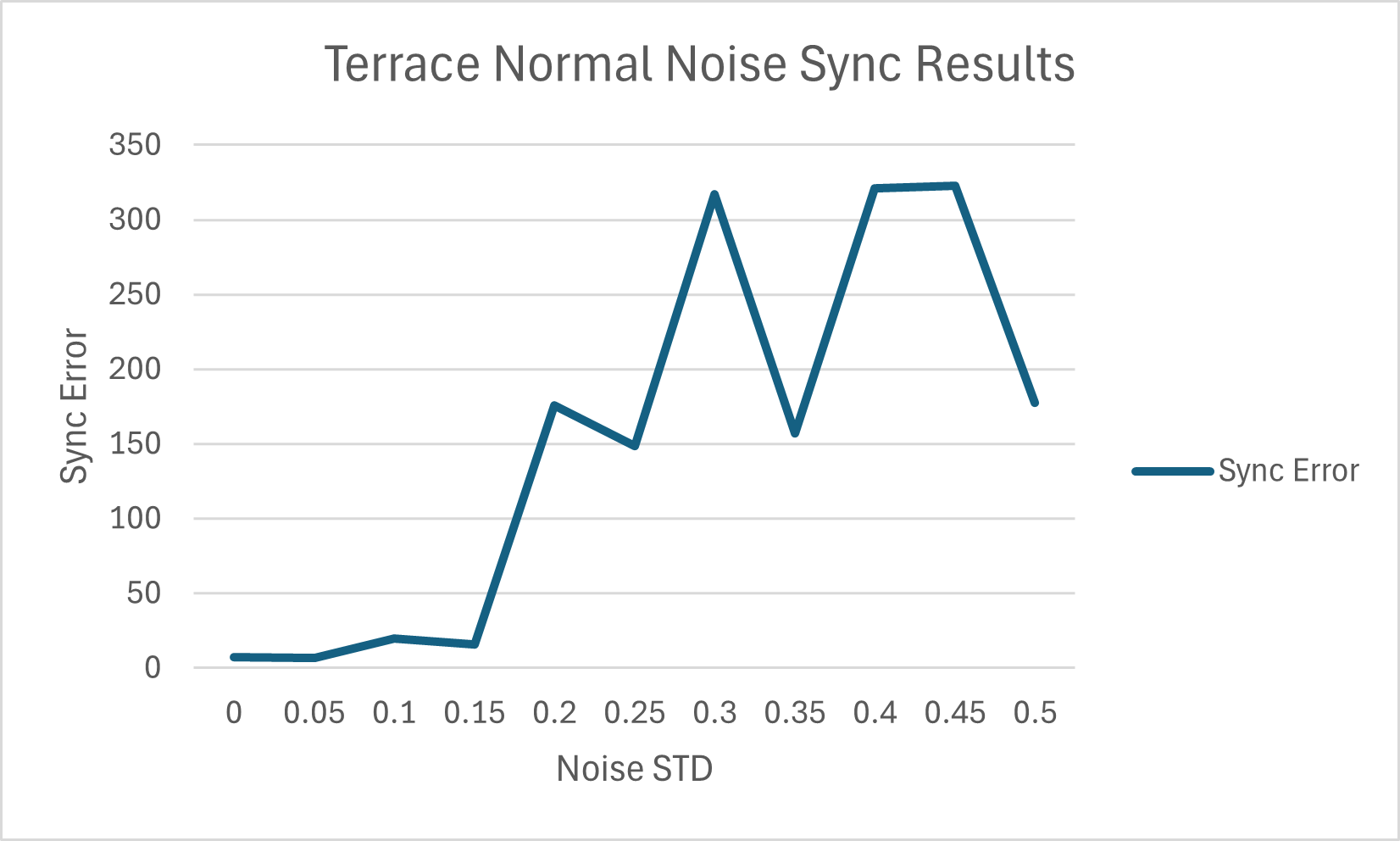}
    \caption{Temporal synchronization error.}
    \label{fig:Terrace_Normal_Noise_Sync_Results}
  \end{subfigure}

  \vspace{1em} 

  \begin{subfigure}{0.5\textwidth}
    \centering
    \includegraphics[width=0.8\linewidth]{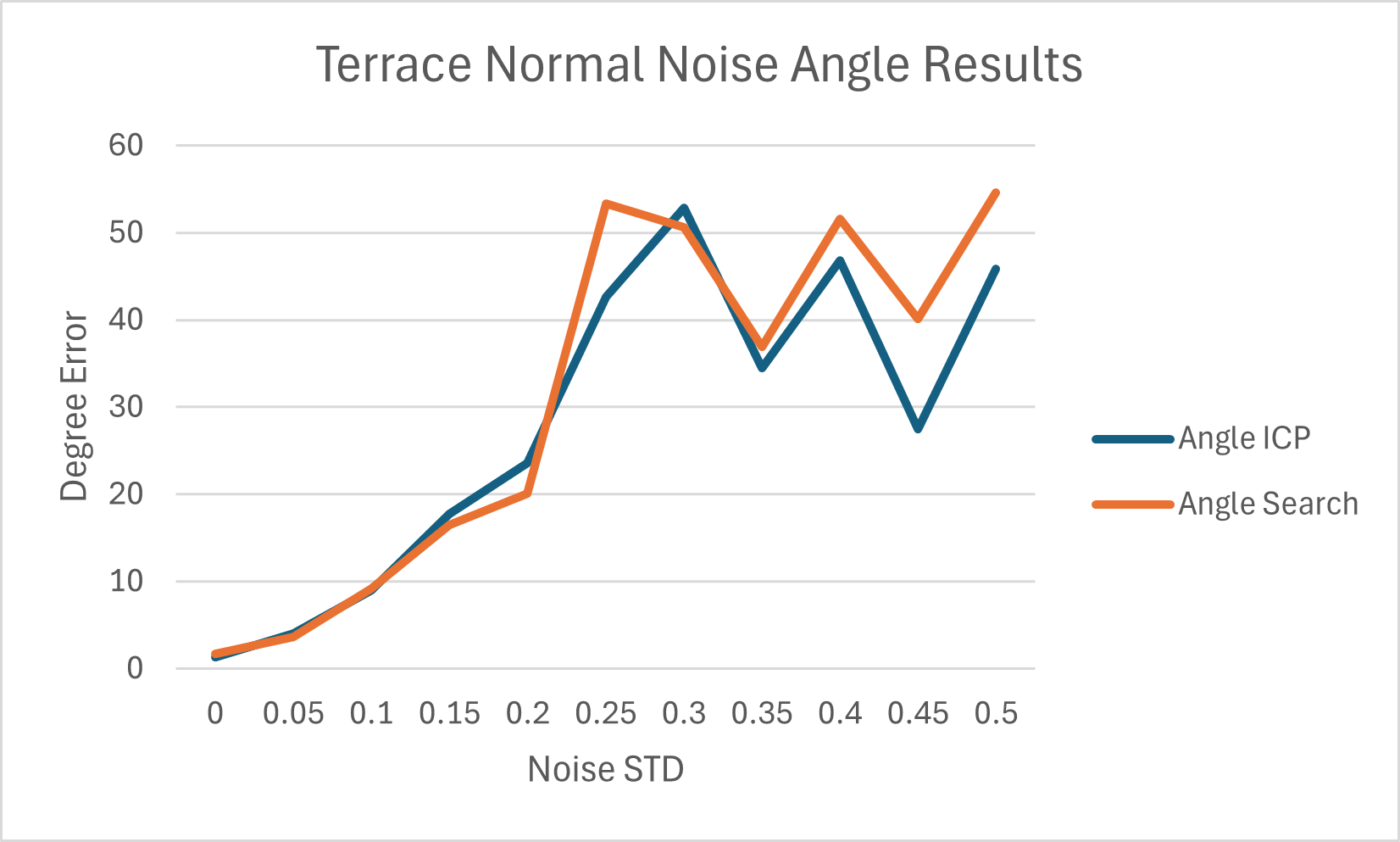}
    \caption{Degree rotation angle error.}
    \label{fig:Terrace_Normal_Noise_Angle_Results}
  \end{subfigure}
  \hfill
  \begin{subfigure}{0.5\textwidth}
    \centering
    \includegraphics[width=0.8\linewidth]{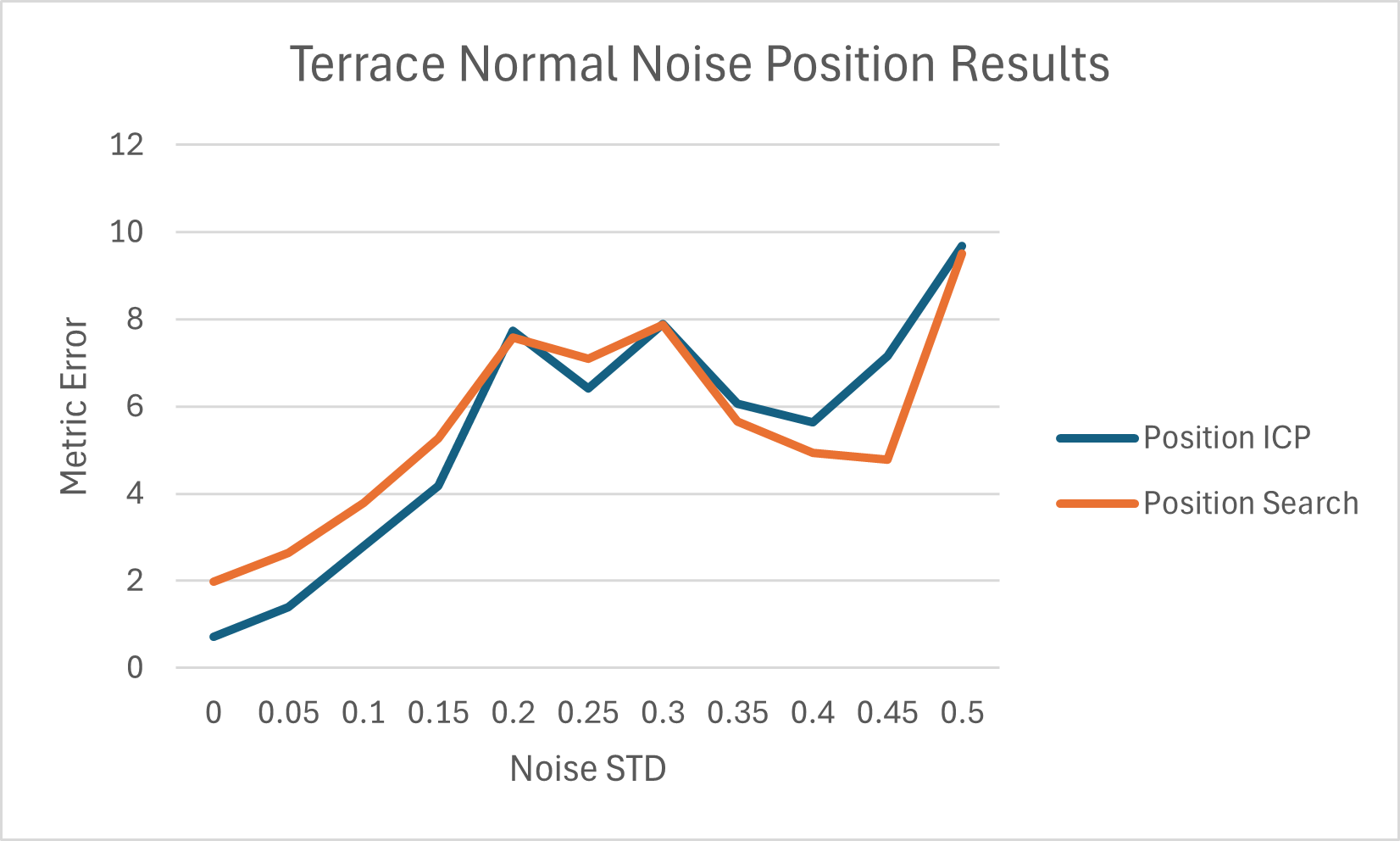}
    \caption{Normalized position error.}
    \label{fig:Terrace_Normal_Noise_Position_Results}
  \end{subfigure}

  \caption{\textbf{Terrace normal noise error curves.}  We plot the error on all of our metrics compared to the standard deviation of the noise added to the normal vector.}
  \label{fig:Terrace_Normal_Noise}
\end{figure}

\paragraph{Normal Noise}
For these experiments, we add random noise to the ground truth normal vector sampled from a Gaussian with mean zero and standard deviation between 0 and 0.5. After adding the noise, we normalize the vector again, run the test for each standard deviation 5 times, and take the median result. 
Fig. \ref{fig:Terrace_Normal_Noise} shows that the higher the offset we add, the worse the rotation, translation, and temporal synchronization results get for the following steps. We see that the greatest increase in error occurs at 0.15 standard deviations.

\begin{figure}
  \centering

  \begin{subfigure}{0.5\textwidth}
    \centering
    \includegraphics[width=0.8\linewidth]{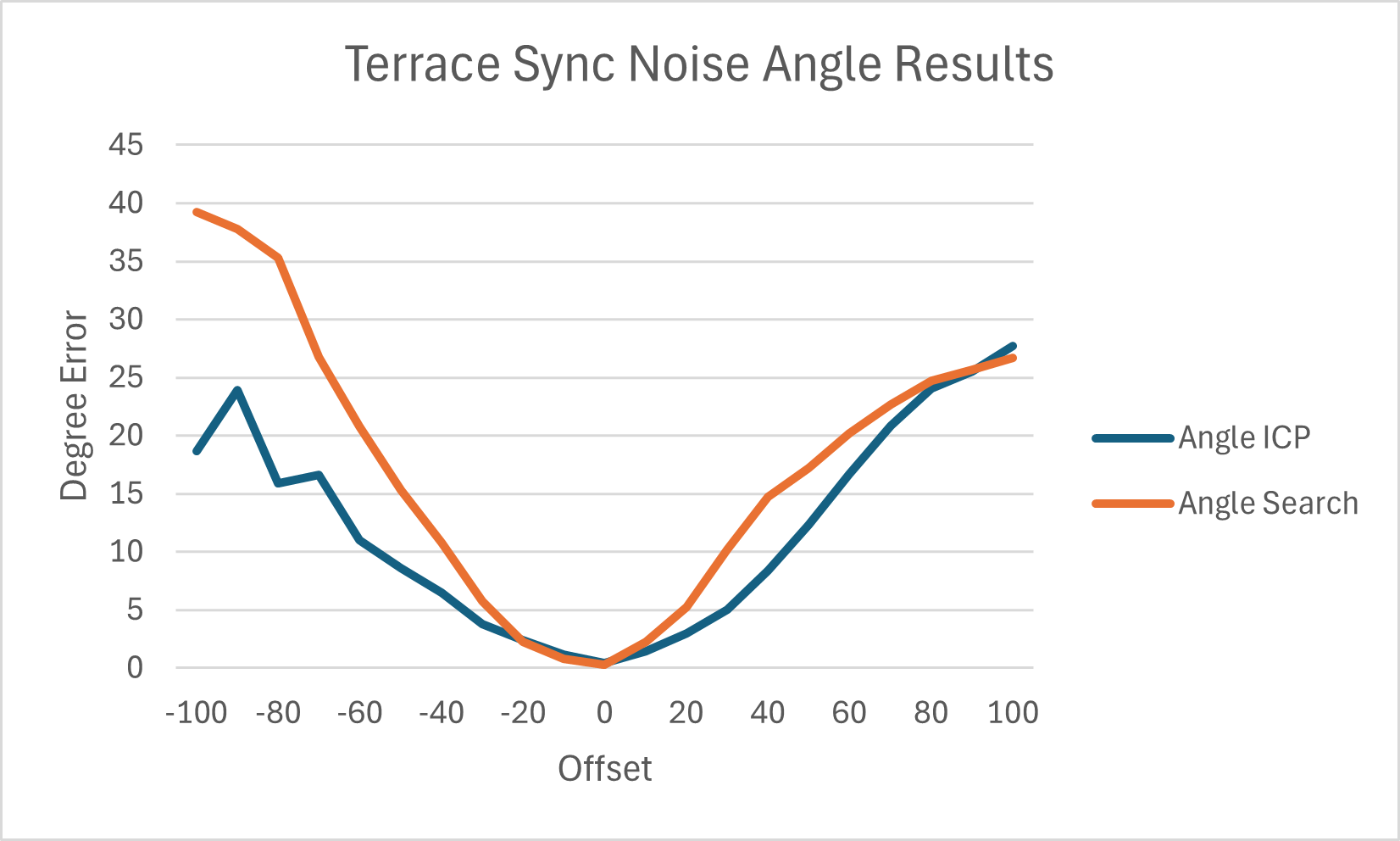}
    \caption{Degree rotation angle error.}
    \label{fig:Terrace_Sync_Noise_Angle_Results}
  \end{subfigure}

  \vspace{1em} 

  \begin{subfigure}{0.5\textwidth}
    \centering
    \includegraphics[width=0.8\linewidth]{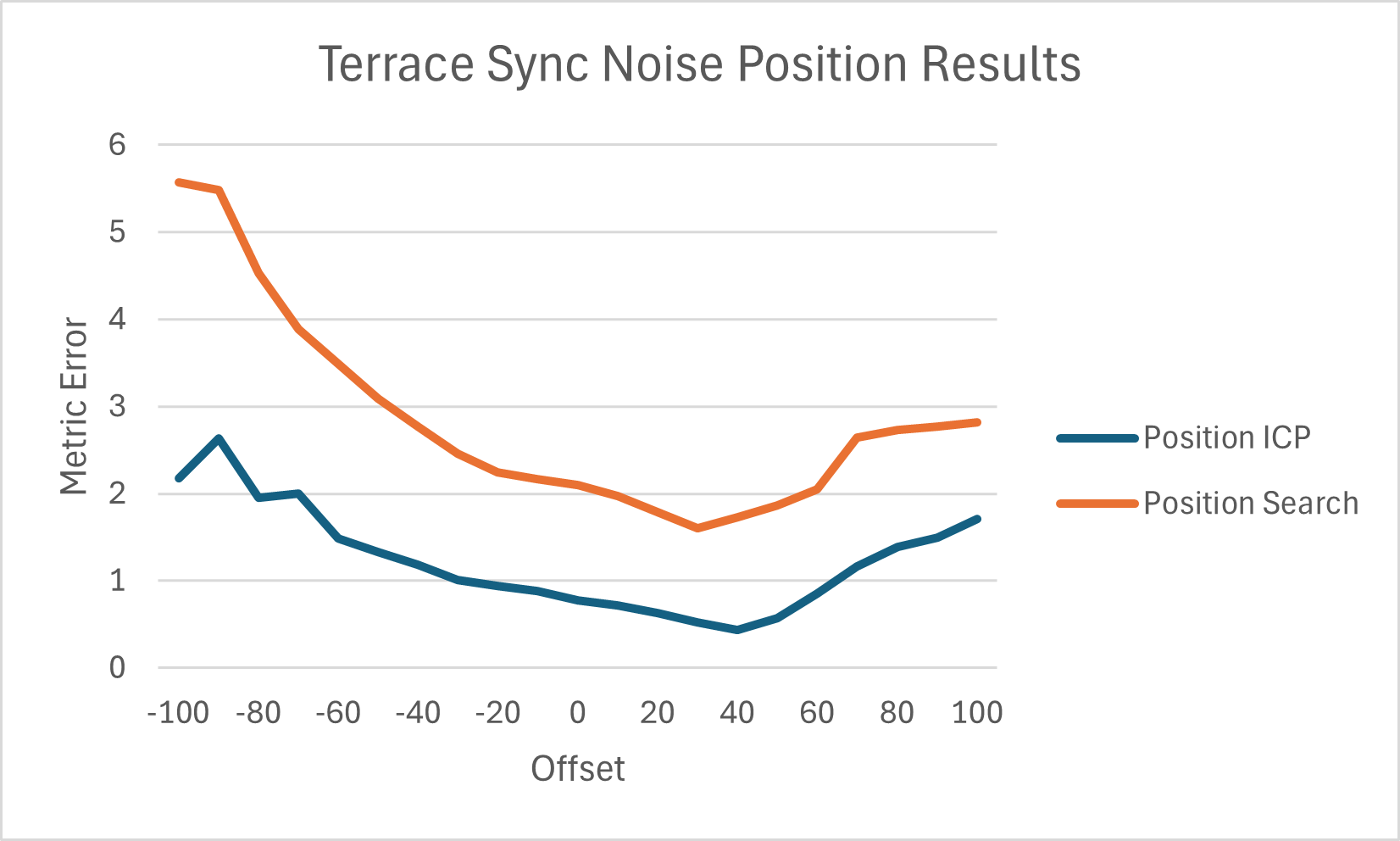}
    \caption{Normalized position error.}
    \label{fig:Terrace_Sync_Noise_Position_Results}
  \end{subfigure}

  \caption{\textbf{Terrace temporal synchronization noise error curves}. We plot the error on all of our metrics compared to the offset added to the temporal synchronization.}
  \label{fig:Terrace_Sync_Noise}
\end{figure}

\paragraph{Time Synchronization Noise}
For these experiments, we add offsets to the temporal synchronization in a range from -100 to 100 frames while using ground truth focal length and normal vector. 
Fig. \ref{fig:Terrace_Sync_Noise} shows the results. The higher the magnitude of offsets we add, the worse the rotation, translation, and temporal synchronization results get for the following steps. We report metrics for all the subsequent steps, enabling a conclusion of how susceptible earlier and later modules are to noise and how this noise propagates through the pipeline.

\begin{figure*}[htbp]
  \centering
  \begin{subfigure}[b]{0.245\textwidth}
    \includegraphics[width=\textwidth]{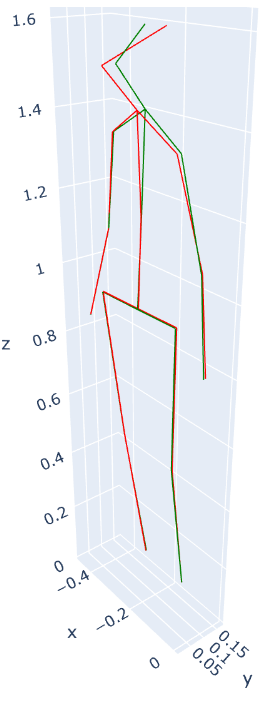}
    \caption{0 degree rotation noise.}
    \label{fig:Human3.6M_Pose_Results1}
  \end{subfigure}
  \begin{subfigure}[b]{0.25\textwidth}
    \includegraphics[width=\textwidth]{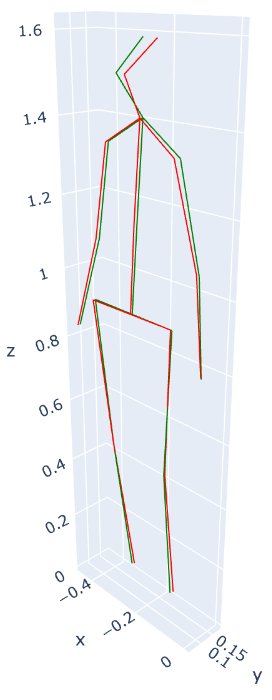}
    \caption{5 degree rotation noise.}
    \label{fig:Human3.6M_Pose_Results2}
  \end{subfigure}

  \caption{\textbf{Reconstructed poses for rotational noise.} Using Subject 1 from Human3.6M on the walking sequence. The green pose represents the ground truth 3D and the red pose represents the reconstructed pose.}
  \label{fig:Human3.6M_Pose_Noise}
\end{figure*}

\begin{figure}
\includegraphics[width=1.0\linewidth]{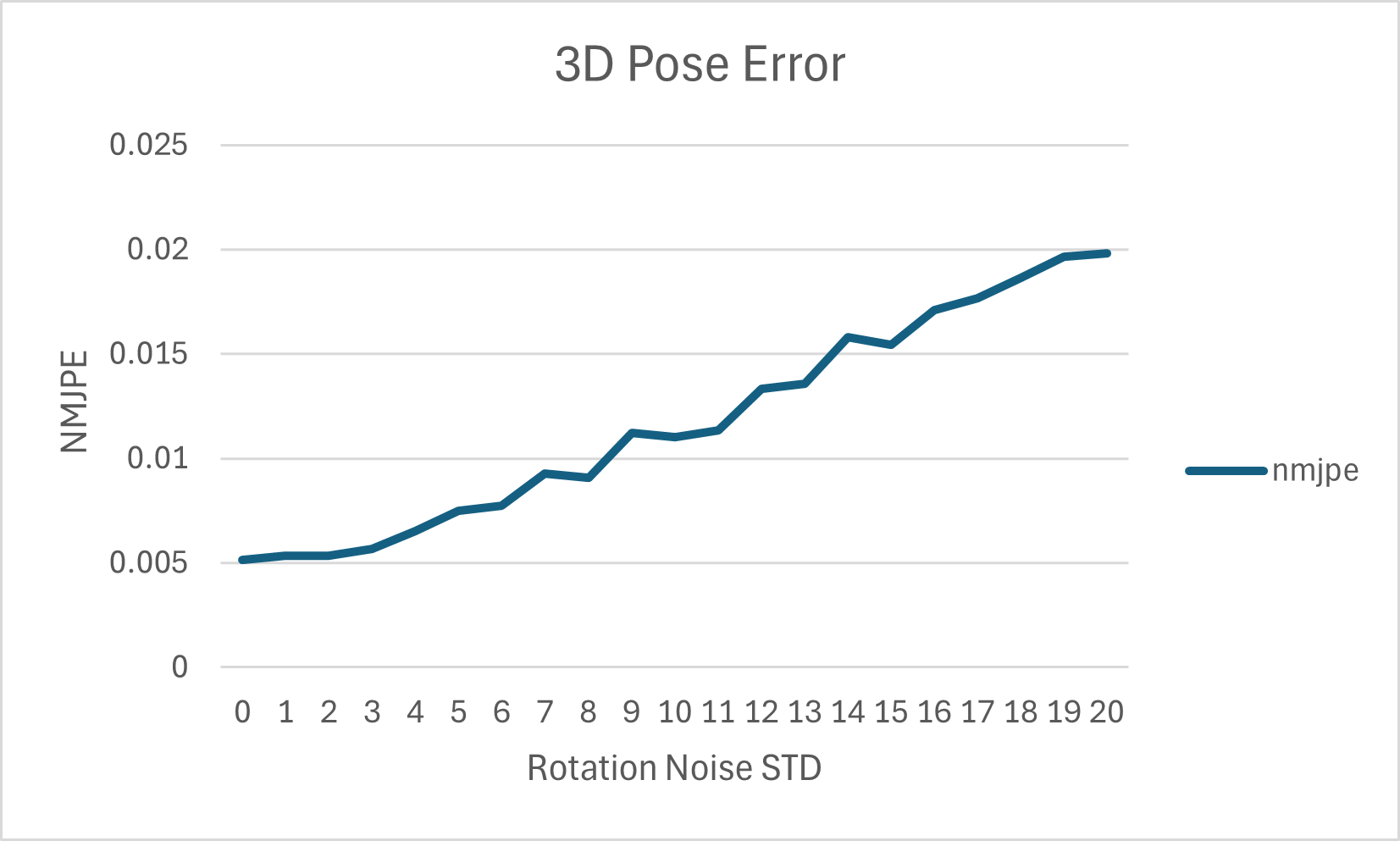}
    \caption{NMPJPE error on the Human3.6M dataset.}
    \label{fig:Human3.6M_Pose_Results_Curve}
\end{figure}

\paragraph{Rotation Noise}
In order to test the effects of the rotation noise, we use Human3.6M to compare the normalized mean per joint position error (NMPJPE) since Human3.6M provides ground truth 3D keypoints. For these experiments, we add random noise to the ground truth rotation matrices sampled from a Gaussian with mean zero and standard deviation between 0 and 20 degrees. We run the test for each standard deviation 5 times and take the median result. 
Fig. \ref{fig:Human3.6M_Pose_Results_Curve} shows that the higher the standard deviation of the rotation noise, the worse the NMPJPE. We see that the error curve is relatively flat from 1 to 3 but starts climbing once we get past 5. However, our qualitative results on subject 1 on the walking sequence are shown in \ref{fig:Human3.6M_Pose_Noise}, we see that the 0 noise case and the 5-degree standard deviation noise look almost identical.

\end{document}